%% file: main.tex
\documentclass{article}
\usepackage{microtype}
\usepackage{graphicx}
\usepackage{subfigure}
\usepackage{booktabs} 
\usepackage{url}
\usepackage{fullpage}
\usepackage{amsmath}
\usepackage{amssymb}
\usepackage{amsthm}
\usepackage{algorithm}
\usepackage{algorithmic}
\usepackage[latin1]{inputenc}
\usepackage{tikz}
\usepackage{pgfplots}
\usepackage{hyperref}
\usetikzlibrary{shapes,arrows}
\usetikzlibrary{positioning}

\input{math_commands}

\tikzstyle{decision} = [diamond, draw, fill=blue!20, 
    text width=4.5em, text badly centered, node distance=3cm, inner sep=0pt]
\tikzstyle{block} = [rectangle, draw, fill=red!20, 
    text width=1em, text centered,  minimum height=4em, scale=0.7]
\tikzstyle{trap} = [trapezium, draw, fill=red!50, 
    text width=1em, text centered, minimum height=4em, shape border rotate=90, scale=.65]  %trapezium left angle=120pt, trapezium right angle=120pt
\tikzstyle{line} = [draw, -latex', line width=0.3mm]
\tikzstyle{cloud} = [draw, circle,fill=red!50, node distance=2cm,
    minimum height=2em]
\tikzstyle{trian} = [regular polygon, regular polygon sides=3, fill=blue!30, shape border rotate=-90,
    text width=1em, text centered, scale=1.3]

\title{Learning Generative Models of Structured Signals from Their Superposition Using GANs with Application to Denoising and Demixing}
\author{Mohammadreza Soltani \\ \href{mailto:msoltani@iastate.edu}{msoltani@iastate.edu} \and Swayambhoo Jain \\ \href{mailto:Swayambhoo.Jain@technicolor.com}{Swayambhoo.Jain@technicolor.com} \and Abhinav Sambasivan \\ \href{mailto:samba014@umn.edu}{samba014@umn.edu}}

\begin{document}
\maketitle

\begin{abstract}
Recently, Generative Adversarial Networks (GANs) have emerged as a popular alternative for modeling complex high dimensional distributions. Most of the existing works implicitly assume that the clean samples from the target distribution are easily available. However, in many applications, this assumption is violated. In this paper, we consider the observation setting when the samples from target distribution are given by the superposition of two structured components and leverage GANs for learning the structure of the components. We propose two novel frameworks: denoising-GAN and demixing-GAN. The denoising-GAN assumes access to clean samples from the second component and try to learn the other distribution, whereas demixing-GAN learns the distribution of the components at the same time. Through extensive numerical experiments, we demonstrate that proposed frameworks can generate clean samples from unknown distributions, and provide competitive performance in tasks such as denoising, demixing, and compressive sensing.
\end{abstract}

\input{./intro}
\input{./prior}
\input{./discuss}
\input{./res}
\input{./append}

\newpage
\bibliographystyle{apalike}  
\bibliography{mrsbiblio}

\end{document}

%% file: math_commands.tex
\usepackage{amsmath,amsfonts,bm}
\usepackage{caption}
\usepackage{multirow}

{
\def\eqref#1{equation~\ref{#1}}
\def\1{\bm{1}}

\DeclareMathAlphabet{\mathsfit}{\encodingdefault}{\sfdefault}{m}{sl}
\SetMathAlphabet{\mathsfit}{bold}{\encodingdefault}{\sfdefault}{bx}{n}

\newcommand{\E}{\mathbb{E}}

\newcommand{\R}{\mathbb{R}}

\def\R{{\mathbb{R}}}

\def\D{{\mathcal{D}}}
\def\X{\mathcal{X}}
\def\N{\mathcal{N}}

\def\'{\prime}

\DeclareMathOperator*{\argmin}{arg\,min}

%% file: intro.tex
\section{Introduction}
\label{intro}
In this paper, we consider the classical problem of separating two structured signals observed under the following superposition model: 
\begin{align} \label{main:eq}
Y = X + N,
\end{align}
where $X \in \X$ and $N \in \N$ are the \emph{constituent signals/components}, and $\X,\N\subseteq\R^p$ denote the two structured sets. In general the separation problem is inherently ill-posed; however, with enough structural assumption on $\X$ and $\N$, it has been established that separation is possible. Depending on the application one might be interested in estimating only $X$ (in this case, $N$ is considered as the corruption), which is referred to as \emph{denoising}, or in recovering both $X$ and $N$ which is referred to as \emph{demixing}. Both demixing and denoising arise in a variety of important practical applications in the areas of signal/image processing, computer vision, machine learning, and statistics~\cite{chen2001atomic, elad2005simultaneous, bobin2007morphological, candes2011rpca}.  

Most of the existing techniques assume some prior knowledge on the structures of $\X$ and $\N$ in order to recover the desired component signal(s). Prior knowledge about the structure of $\X$ and $\N$ can only be obtained if one has access to the generative mechanism of the signals or has access to clean samples from the probability distribution defined over sets $\X$ and $\N$. In many practical settings, neither of these may be feasible. In this paper, we consider the problem of separating constituent signals from superposed observations when clean access to samples from the distribution is not available. In particular, we are given a set of superposed observations $\{Y_i = X_i + N_i\}_{i=1}^K$ where $X_i \in \X$ and $Y_i \in \N$ ($\X$ and $\N$ are not known in general) are i.i.d samples from their respective (unknowns) distributions. In this setup, we explore two questions: First, \emph{How can one learn prior knowledge about the individual components from superposition samples?} Second, \emph{Can we leverage the implicitly learned constituent distributions for tasks such as denoising and demixing?}

\subsection{Setup and Our Technique}

%Estimating of constituent components from their superposition inherently suffers from an ambiguity problem. It is well known that in order to obtain the components, the underlying components should have some structure such that they are distinguishable enough from each other~\citep{soltani2016fastIEEETSP17, soltani2017iterative, druce2016defect}. 
%In this paper, we assume that both components $X$ and $N$ have some structure. 
Motivated by the recent success of generative models in high dimensional statistical inference tasks such as compressed sensing in \cite{bora2017compressed, bora2018ambientgan}, in this paper, we focus on Generative Adversarial Network (GAN) based generative models to implicitly learn the distributions, i.e., generate samples from their distributions. Most of the existing works on GANs typically assume access to clean samples from the underlying signal distribution. However, this assumption clearly breaks down in the superposition model considered in our setup, where the structured superposition makes training generative models very challenging. 

In this context, we investigate the first question with varying degrees of assumption about the access to clean samples from the two signal sources. We first focus on the setting when we have access to samples only from the constituent signal class $\N$ and observations, $Y_i$'s. In this regard, we propose the \emph{denoising}-GAN framework. 
% training a GAN, given enough number of  and samples from only the constituent signal class $\N$, using what we call . 
However, assuming access to samples from one of the constituent signal class can be restrictive and is often not feasible in real-world applications. Hence, we further relax this assumption and consider the more challenging demixing problem, where samples from the second constituent component are not available and solve it using what we call the \emph{demixing}-GAN framework. 

Finally, to answer the second question, we use our trained generator(s) from the proposed GAN frameworks for denoising and demixing tasks on unseen test samples (i.e., samples not used in the training process) by discovering the best hidden representation of the constituent components from the generative models. In addition to the denoising and demixing problems, we also consider a compressive sensing setting to test the trained generator(s). {Below we explicitly list the contribution made in this paper:} 
% In this regard, we study the performance of our previously introduced denoising and demixing framework along with GAN in comparison with the vanilla GAN. 
\begin{enumerate}
	\item Under the assumption that one has access to the samples from one of the constituent component, we extend the canonical GAN framework and propose \emph{denoising}-GAN framework. This learns the prior from the training data that is heavily corrupted by additive structured component. We demonstrate its utility in denoising task via numerical experiments.
	\item We extend the above denoising-GAN and propose \emph{demixing}-GAN framework. This learns the prior for both the constituent components from their superpositions, without access to separate samples from any of the individual components. We demonstrate its utility in demixing task via numerical experiments.
	%\item We provide empirical evidence for the quality of the generative models trained compressed sensing in structured corruption for natural images. 
\end{enumerate}  

The rest of this paper is organized as follows: In section $2$, we discuss relevant previous works and compare our novelty over these existing methods. In section $3$, we formally introduce our proposed approach, and in section $4$, we provide some experimental results to validate our framework. Finally, we conclude this paper by providing a summary of the paper in section $5$.

%% file: prior.tex
\section{Application and Prior Art}
\label{prior}
To overcome the inherent ambiguity issue in problem~(\ref{main:eq}), many existing methods have assumed that the structures of sets (i.e., the structures can be low-rank matrices, or have sparse representation in some domain~\cite{mccoyTropp2014}) $\X$ and $\N$ are a prior known and also that the signals from $\X$ and $\N$ are ``distinguishable" ~\cite{elad2006image, soltani2017iterative, soltani2016fastIEEETSP17, druce2016defect, elyaderani2017group, jain2017compressed}. The assumption of having the prior knowledge is a big restriction in many real-world applications. Recently, there have been some attempts to automate this \emph{hard-coding} approach. Among them, structured sparsity~\cite{hegde2015approximation}, dictionary learning~\cite{elad2006image}, and in general manifold learning are the prominent ones. While these approaches have been successful to some extent, they still cannot fully address the need for the prior structure.  
%On the other hand, the re-emergence of neural networks (deep learning) and its success in numerous applications has proved it as a very powerful tool to this end. 
Over the last decade, deep neural networks have been demonstrated to learn useful representations of real-world signals such as natural images, and thus have helped us understand the structure of the high dimensional signals, for e.g. using deep generative models ~\cite{ulyanov2017deep}.

In this paper, we focus on Generative Adversarial Networks (GANs)~\cite{goodfellow2014generative} as the generative models for implicitly learning the distribution of constituent components. GANs have been established as a very successful tool for generating structured high-dimensional signals~\cite{berthelot2017began, vondrick2016generating} as they do not directly learn a probability distribution; instead, they generate samples from the target distribution(s)~\cite{goodfellow2016nips}. In particular, if we assume that the structured signals are drawn from a distribution lying on a low-dimensional manifold, GANs generate points in the high-dimensional space that resemble those coming from the true underlying distribution. 
 
Since their inception by \cite{goodfellow2014generative}, there has been a flurry of works on GANs~\cite{zhu2017unpaired, yeh2016semantic, subakan2018generative} to name a few. In most of the existing works on GANs with few notable exceptions \cite{wu2016learning, bora2018ambientgan, kabkab2018task, hand2018phase, zhu2016generative}, it is implicitly assumed that one has access to clean samples of the desired signal. However, in many practical scenarios, the desired signal is often accompanied by unnecessary components. Recently, GANs have also been used for capturing of the structure of high-dimensional signals specifically for solving inverse problems such as sparse recovery, compressive sensing, and phase retrieval~\cite{bora2017compressed, kabkab2018task, hand2018phase}. Specifically, \cite{bora2017compressed} have shown that generative models provide a good prior to structured signals, for e.g., natural images, under compressive sensing settings over sparsity-based recovery methods. They rigorously analyze the statistical properties of a generative model based on compressed sensing and provide theoretical guarantees and experimental evidence to support their claims. However, they don't explicitly propose an optimization procedure to solve the recovery problem. They simply suggest using stochastic gradient-based methods in the low-dimensional latent space to recover the signal of interest. This has been addressed by \cite{shah2018solving}, where the authors propose using a projected gradient descent algorithm for solving the recovery problem directly in the ambient space (space of the desired signal). They provide theoretical guarantees for the convergence of their algorithm and also demonstrate improved empirical results over~\cite{bora2017compressed}.% under these settings. 
 
While GANs have found many applications, most of them need direct access to the clean samples from the unknown distribution, which is not the case in many real applications such as medical imaging. AmbientGAN framework~\cite{bora2018ambientgan} partially addresses this problem. In particular, they studied various measurement models and showed that their GAN can find samples of clean signals from corrupted observations. Although similar to our effort, there are several key differences between ours and AmbientGAN. Firstly, AmbientGAN assumes that the measurement model and parameters are known, which is a very strong and limiting assumption in real applications. One of our main contributions is addressing this limitation by studying the demixing problem. Second, for the noisy measurement settings, AmbientGAN assumes an arbitrary measurement noise and no corruption in the underlying component. We consider the corruption models in the signal domain rather than as a measurement one. This allows us to study the denoising problem from a highly structured corruption. Lastly, their approach just learns the distribution of the clean images; however, it has not been used for the task of image denoisng (i.e., how to denoise an unseen corrupted image). Our framework addresses this issue as well.

%1- An Unsupervised Approach to Solving Inverse Problems using Generative Adversarial Networks, Rushil Anirudh, et al,\\

%% file: discuss.tex
\section{Background and the Proposed Idea}
\label{discuss}

\subsection{Background}
Generative Adversarial Networks (GANs) are one of the successful generative models in practice was first introduced by~\cite{goodfellow2014generative} for generating samples from an unknown target distribution. As opposed to the other approaches for density estimation such as \emph{Variational Auto-Encoders (VAEs)}~\cite{kingma2013auto}, which try to learn the distribution itself, GANs are designed to generate samples from the target probability density function. This is done through a zero-sum game between two players, \emph{generator}, $G$ and \emph{discriminator}, $D$ in which the generator  $G$ plays the role of producing the fake samples and discriminator $D$ plays the role of a cop to find the fake and genuine samples. Mathematically, this is accomplished through the following \emph{min-max} optimization problem:
\begin{align}
\label{GANprob}
\min_{\theta_g}\max_{\theta_d} ~~~\E_{x\thicksim\D_x}[log(D_{\theta_d}(x))] \E_{z\thicksim\D_z}[log(1-D_{\theta_d}(G_{\theta_g}(z)))],
\end{align}
where $\theta_g$ and $\theta_d$ are the parameters of generator networks and discriminator network respectively,  and $\D_x$ denotes the target probability distribution , and $\D_z$ represents the probability distribution of the hidden variables $z\in\R^h$, which is assumed either a uniform distribution in $[-1,1]^h$, or standard normal. One can also use identity function instead of $log(.)$ function in the above expression. The resulting formulation is called WGAN~\cite{pmlr-v70-arjovsky17a}. It has been shown that if $G$ and $D$ have enough capacity, then solving optimization problem~(\ref{GANprob}) by alternative stochastic gradient descent algorithm guarantees the distribution  $\D_g$  at the output of the generator converges to $\D_x$. Having discussed the basic setup of GANs, next we present the proposed modifications to the basic GAN setup that allows for usage of GANs as a generative model for denoising and demixing structured signals. 

\begin{figure*}   
\centering
\begin{tabular}{cccc}
\begin{tikzpicture}[scale=.15, node distance = 1.3cm, auto]
\node (gen) [trap] {$G_{\theta_g}$};
\node (Z) [block, left of=gen, fill=red!50] {Z};
\path [line] (Z) -- (gen);
\coordinate (K) at (24, 14);
\node(clean) [inner sep=0pt, left =1.5cm of K] {\includegraphics [width=0.08\linewidth]{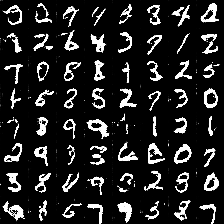}};
\path [line] (7.5,0) -- node{} (7.5,10);
\coordinate (D) at (0.57,0);
\node(noise) [inner sep=0pt, left =1.7cm of D] {\includegraphics [width=0.04\linewidth]{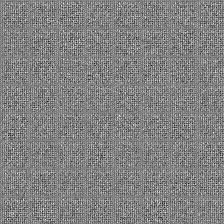}};
\path [line] (noise) -- (Z);
\node (add) [cloud, right of=gen] {$+$}; 
\path [line] (gen) -- (add);
\node (corrup) [block, above of=add, fill=yellow!100, minimum height=1em, minimum width=2em, text width=4em] {$N\in\N$}; 
\path [line] (corrup) -- (add);
\coordinate (A) at (0.1,-1.75);
\node (Disc) [trian, right = 3cm of A, minimum height=2em, minimum width=4em] {$D_{\theta_d}$}; 
\coordinate (B) at (7,-3.5);
\path [line] (add)  edge[out=5, in=170] (Disc);         
\node (dataset) [block,  fill=green!50, below = 6ex of B, minimum height=4em, minimum width=8em, text width=8em, rounded corners] {Observed Samples\\$\{y_1,y_2,\ldots,y_n\}$};
\path [line] (dataset) edge[out=-5, in=170]  (Disc);
\coordinate (C) at (-0.5,-14.3);
\node(obs) [inner sep=0pt, left =.5cm of C] {\includegraphics [width=0.08\linewidth]{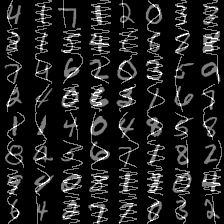}};
\path [line] (obs) -- (dataset);
\path [line] (Disc) --  node [above, midway] {\tiny{0: fake}} node [below, midway] {\tiny{1: genuine}} (40,-1.73);
\end{tikzpicture}& & 

\begin{tikzpicture}[scale=.7, transform shape, node distance = 1.7cm, auto]
\coordinate (AD) at (0.5,-1.5);
\node (gen1) [trap] {$G_{\theta_{g_1}}$};
\node (Z1) [block, left of=gen1, fill=red!50] {$Z_1$};
\path [line] (Z1) -- (gen1);
\node (add) [cloud, right = 1cm of AD] {$+$}; 
\path [line] (gen1) edge[out=0, in=-175] (add);
\node (gen) [trap, below = 1cm of gen1] {$G_{\theta_{g_2}}$};
\node (Z) [block, left of=gen, fill=red!50] {$Z_2$};
\path [line] (Z) -- (gen);
\node (add) [cloud, right = 1cm of AD] {$+$}; 
\path [line] (gen)edge[out=0, in=185](add);
\coordinate (K) at (4.7,-3.4);
\node(clean) [inner sep=0pt, left =2cm of K] {\includegraphics [width=0.115\linewidth]{./All_talks/gen1_sin/x_sample_55.png}};
\path [line] (1.05,-2) -- node{} (2,-2.75);
\coordinate (E) at (-0.75,.5);
\node(Cor) [inner sep=0pt, right =1.7cm of E] {\includegraphics [width=0.115\linewidth]{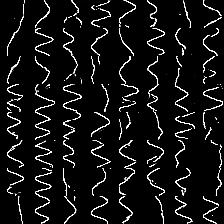}};
\path [line] (1.05,-.9) -- node{} (2,-0.1);
\coordinate (A) at (0.5,-1.54);
\node (Disc) [trian, right = 3.5cm of A, minimum height=2em, minimum width=4em] {$D_{\theta_d}$}; 
\coordinate (B) at (0.05,-5);
\path [line] (add) -- (Disc);           
\node (dataset) [block,  fill=green!50, below = 1ex of B, minimum height=4em, minimum width=8em, text width=8em, rounded corners] {Observed Samples\\$\{y_1,y_2,\ldots,y_n\}$};
\path [line] (dataset) edge[out=-8, in=218]  (Disc);
\path [line] (Disc) -- node [above, midway] {0: fake} node [below, midway] {1: genuine} (6.5,-1.55);
\coordinate (C) at (-0.5,-5.65);
\node(obs) [inner sep=0pt, left =2cm of C] {\includegraphics [width=0.13\linewidth]{corrup_64_sin1.png}};
\path [line] (obs) -- (dataset);
\coordinate (D) at (0.57,0);
\node(noise) [inner sep=0pt, left =3cm of D] {\includegraphics [width=0.07\linewidth]{./All_talks/noise1.png}};
\path [line] (noise) -- (Z1);
\coordinate (D1) at (0.57,-2.5);
\node(noise1) [inner sep=0pt, left =3cm of D1] {\includegraphics [width=0.07\linewidth]{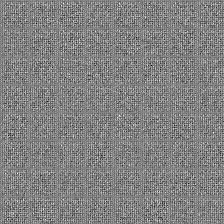}};
\path [line] (noise1) -- (Z);
\end{tikzpicture}\\
& & \\
(a)&  & (b) 
\end{tabular}
\caption{\small{The architecture of proposed GANs. (a) denoising-GAN. (b) demixing-GAN.}}
\label{arcdonisedemix}
\end{figure*}

\subsection{denoising-GAN}
Our idea is inspired by AmbientGAN due to~\cite{bora2018ambientgan} in which they used a regular GAN architecture to solve some inverse problems such as inpainting, denoising from unstructured noise, and so on. In particular, assume that instead of clean samples, one has access to a corrupted version of the samples where the corruption model is captured by a known random function $f_{{N}}(n)$ where ${N}$ is a random variable. For instance, the corrupted samples can be generated via just adding noise, i.e., $Y = X + {N}$. The idea is to feed discriminator with observed samples, $y_i$'s (distributed as $Y$) together with the output of generator $G$ which is corrupted by model $f_{{N}}(n)$.
Our denoising-GAN framework is illustrated in Figure~\ref{arcdonisedemix}(a). This framework is similar to one proposed in Ambient GAN paper; however, the authors did not consider denoising task from structured superposition based corruption.  In the experiment section, we show that denoising-GAN framework can generate clean samples even from structured corruption.   
Now we use the trained denoising-GAN framework for denoising of a new test corrupted image which has not been used in the training process. To this end, we use our assumption that the components have some structure and the representation of this structure is given by the last layer of the trained generator,  i.e., $X\in G_{\widehat{\theta}_g}$\footnote{$G_{\widehat{\theta}_g}(.)$ denotes the trained generator network with parameter $\widehat{\theta}_g$.}. This observation together with this fact that in GANs, the low-dimension random vector $z$ is representing the hidden variables, leads us to this point: in order to denoise a new test image, we have to find a hidden representation, giving the smallest distance to the corrupted image in the space of $G_{\widehat{\theta}_g}$~\cite{shah2018solving, bora2017compressed}. In other words, we have to solve the following optimization problem: 
\begin{align}
\label{denoisetest}
\widehat{z} = \argmin_z\|u - G_{\widehat{\theta}_g}(z)\|_2^2 + \lambda\|z\|_2^2,  %{z\in[-1,1]^{100}}
\end{align}
where $u$ denotes the corrupted test image. The solution of this optimization problem provides the (best) hidden representation for an unseen image. Thus, the clean image can be reconstructed by evaluating $G_{\widehat{\theta}_g}(\widehat{z})$. While optimization problem~(\ref{denoisetest}) is non-convex, we can still solve it by running gradient descent algorithm in order to get a stationary point\footnote{While we cannot guarantee the stationary point is a local minimum, but the empirical experiments show that gradient descent (implemented by backpropagation) can provide a good quality result.}.
%While this results a good performance, it requires that the model of corruption to be known even for simple additive noise.

\subsection{Theoretical Insights for the denoising-GAN}
Here, we revisit some theoretical arguments pioneered by \cite{bora2018ambientgan}. As authors have discussed in this paper in Lemma $5.1$ and Theorem $5.2$, so long as there is a bijection map from the probability distribution of observation space ($Y$-domain) to the probability distribution of signal space ($X$-domain), and by choosing optimal discriminator as $D = \frac{\D_y}{\D_y + \D_g}$, then it is guaranteed that the generator, $G$ is optimal if and only if $\D_g = \D_y$. Here, $\D_g$ and $\D_y$ denote the probability distribution of generator and observation (corrupted signal), respectively. This can be proved by arguments given in the original GAN paper by~\cite{goodfellow2014generative}. Authors in~\cite{bora2018ambientgan} showed that the uniqueness map assumption is satisfied in some special cases. Since the probability distribution of the sum is given by convolution, from equation~(\ref{main:eq}), we have: $\D_y = \D_x*\D_n,$ where $\D_N$ denotes the distribution of the corruption part, $N$ which we have access to the samples from it, and $*$ denotes the convolution operator. If we take Fourier transform (or using the characteristic function, $\Phi(.)$) from this equation, we obtain: $\Phi_y = \Phi_x. \Phi_n$. As a result, $\D_x = \Phi^{-1}_x(\frac{\Phi_y}{\Phi_n})$. This means that the probability distribution of signal domain is determined uniquely (due to one-to-one relation between the probability distribution and the characteristic function). For $\D_x = \Phi^{-1}_x(\frac{\Phi_y}{\Phi_n})$ to be well defined a straightforward condition is that the $\Phi_n$ is non-zero almost everywhere. This condition is satisfied for many corruption distributions. For example, consider the case the true signal is distributed as Gaussian, and the corruption samples are drawn from uniform distribution between $[-1,1]$ the $\Phi_n$ is non-zero almost everywhere. This is also satisfied for a $n$-dimensional random sparse vector with $k$ non-zero entries such that there is non-zero probability of each set of $k$-entries and the joint-distribution of non-zero $k$-entries is Gaussian with full-rank covariance matrices. This covers many structured noises.

\subsection{demixing-GAN}
Now, we go through our main contribution, demixing. Figure~\ref{arcdonisedemix}(b) shows the GAN architecture, we are using for the purpose of separating or demixing of two structured signals form their superposition. As illustrated, we have used two generators and have fed them with two random noise vectors $z_1\in\R^{h_1}$ and $z_2\in\R^{h_2}$ according to a uniform distribution defined on a hyper-cube, where $h_1, h_2$ are less than the dimension of the input images. We also assume that they are independent of each other. 
Next, the output of generators are summed up and the result is fed to the discriminator along with the superposition samples, $y_i's$. 
In Figure~\ref{arcdonisedemix}(b), we just show the output of each generator after training for an experiment case in which the mixed image consists of 64 MNIST binary image~\cite{lecun-mnisthandwrittendigit-2010} (for $X$ part) and a second component constructed by random sinusoidal (for $N$ part) (please see the experiment appendix for the details of this specific experiments). Somewhat surprisingly, this architecture based on two generators can produce samples from the distribution of each component after enough number of training iterations. We note that this approach is fully unsupervised as we only have access to the mixed samples and nothing from the samples of constituent components is known. As mentioned above, this is in sharp contrast with AmbientGAN and our previous structured denoising approach. As a result, the demixing-GAN framework can generate samples from the second components (for example random sinusoidal, which further can be used in the task of denoising where the corruption components are sampled from highly structured sinusoidal waves). Now similar to the denoising-GAN framework, we can use the trained generators in Figure~\ref{arcdonisedemix}(b), for demixing of the constituent components for a given test mixed image which has not been used in training. Similarly, we can solve the following optimization problem:
\vspace{-.1cm}
\begin{align}
\label{demixtest}
\widehat{z_1}, \widehat{z_2} = \argmin_{z_1,z_2}\|y - G_{\widehat{\theta}_{g_1}}(z_1) -  G_{\widehat{\theta}_{g_2}}(z_2)\|_2^2+\lambda_1\|z_1\|_2^2++\lambda_2\|z_2\|_2^2,  %_{z_1,z_2\in[-1,1]^{100}}
\end{align}
where $u$ denotes the test mixed image.

Now, each component can be estimated by evaluating $G_{\widehat{\theta}_{g_1}}(\widehat{z_1})$ and $G_{\widehat{\theta}_{g_2}}(\widehat{z_2})$\footnote{$G_{\widehat{\theta}_{g_1}}(.)$ and $G_{\widehat{\theta}_{g_2}}(.)$ denote the first and second trained generator with parameter $\widehat{\theta}_{g_1}$ and $\widehat{\theta}_{g_2}$, respectively.}. Similar to the previous case, while the optimization problem in~(\ref{demixtest}) is non-convex, we can still solve it through block coordinate gradient descent algorithm, or in a alternative minimization fashion. We note that in both optimization problems~(\ref{denoisetest}) and~(\ref{demixtest}), we did not project on the box sets on which $z_1$ and $z_2$ lie on. Instead we have used regularizer terms in the objective functions (which are not meant as projection step). We empirically have observed that imposing these regularizers can help to obtain good quality images in our experiment; plus, they may help that the gradient flow to be close in the region of interest by generators. This is also used in~\cite{bora2017compressed}. Finally, we have provided some theoretical intuitions for the demixing-GAN in the appendix.

%% file: res.tex
\section{Numerical Experiments}
\label{res}

In this section, we present various experiments showing the efficacy of the proposed frameworks (depicted in Figure~\ref{arcdonisedemix}(a) and Figure~\ref{arcdonisedemix}(b)) in three different setups. First, we will focus on the denoising from structured corruption both in training and testing scenarios. Next, we focus on demixing signals from structured distributions. Finally, we explore the use of generative models from the proposed GAN frameworks in compressive sensing setup. In all the following experiments, we did our best for choosing all the hyper-parameters. We defer the details of experiments setup, the complementary experiments on the compressive sensing scenario, and experiments on the other datasets (F-MNIST~\cite{xiao2017}, combination of F-MNIST and MNIST, SVHN~\cite{netzer2011reading}, and Quick-Draw~\cite{QuickDraw}) to the appendix.

\subsection{Structured Corruption Models}
\begin{figure*}[h]
\hspace{.2cm}
\begin{tabular}{ccc}
\includegraphics [width=0.12\linewidth]{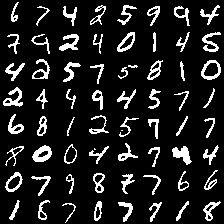}&
\includegraphics [width=0.12\linewidth]{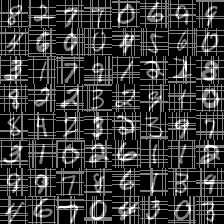}&
\includegraphics [width=0.12\linewidth]{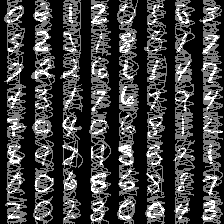} \\
(a) &(b) & (c)
\end{tabular}
\hspace{.5cm}
\begin{tabular}{ccccccccc}
\includegraphics [width=0.1\linewidth]{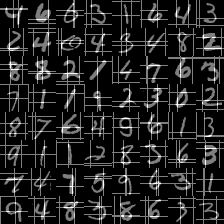}&&&&&
\includegraphics [width=0.1\linewidth]{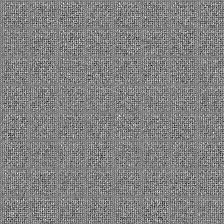}&
\includegraphics [width=0.1\linewidth]{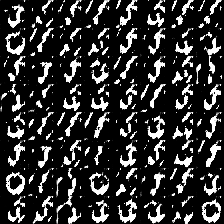}&
\includegraphics [width=0.1\linewidth]{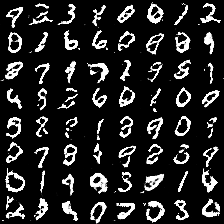} &
\includegraphics [width=0.1\linewidth]{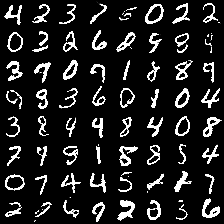} \\
\includegraphics [width=0.1\linewidth]{corrup_64_sin1.png}&&&&&
\includegraphics [width=0.1\linewidth]{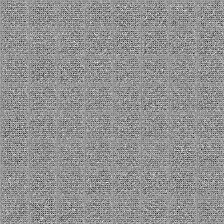}&
\includegraphics [width=0.1\linewidth]{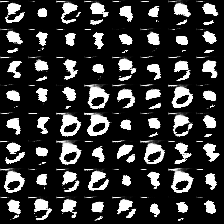}&
\includegraphics [width=0.1\linewidth]{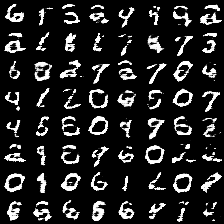}& 
\includegraphics [width=0.1\linewidth]{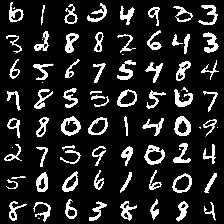}\\
\tiny{Corrupted Input} & &&&& \tiny{$1^{st}$ epoch} & \tiny{$2^{nd}$ epoch} & \tiny{$5^{th}$ epoch} & \tiny{$64^{th}$ epoch}
\end{tabular}
%\caption{}
\caption{\small{Left panel: (a). Clean binary MNIST image. (b). Corrupted image with random horizontal and vertical lines. (C). Corrupted image with random sinusoidal waves. Right panel: Evolution of outputted samples by generator for fixed $z$. Top row is for random horizontal and vertical corruption. Bottom row is for random sinusoidal corruption.}}
\label{corruptionTrueMNIS}
\end{figure*}

For all the experiments in this section, we have used the network architectures for discriminator and generator(s) similar to the one proposed in DCGAN~\cite{radford2015unsupervised}. DCGAN is a CNN based GAN consists of convolutional layers followed by batch normalization (except the last layer of the generator and first layer of discriminator). We have also considered the binary MNIST dataset as the clean ground-truth component. For the corruption part, we have used two structured noise models similar to~\cite{chen2014removing}. In the first one, we generate random vertical and horizontal lines and add them to the dataset. The second structured noise is constructed based on random sinusoidal waves in which the amplitude, frequency, and phase are random numbers. We define the level of corruption (lc) as the number of sinusoidal or lines added to the original image. We note that both of these corruption models are highly structured. These two corruption models along with the clean MNIST images have been shown in the left panel of Figure~\ref{corruptionTrueMNIS}. 
 
\subsection{Denoising from Structured Corruption -- Training}   %such that each epoch includes $937$ batches
In this section, we use GAN architecture illustrated in Figure~\ref{arcdonisedemix}(a) for removing of the structured noise. The setup of the experiment is as follows: we use $55000$ images with size $28\times 28$ corrupted by either of the above corruption models\footnote{We set level of corruption $1$ for sinusoidal and $2$ for the vertical and horizontal lines in the training of the denoising-GAN.}. The resulting images, $y_i$'s are fed to the discriminator. We also use hidden random vector $z\in\R^{100}$ drawn from a uniform distribution in $[-1,1]^{100}$ for the input of the generator. During the training, we use $64$ mini-batches along with the regular loss function in the GAN literature, stated in problem~(\ref{GANprob}). The optimization algorithm is set to Adam optimizer, and we train discriminator and generator one time in each iteration. We set the number of epochs to $64$. To show the evolution of the quality of output samples by the generator, we fix an input vector $z$ and save the output of the generator at different times during the training process. 
The right panel of Figure~\ref{corruptionTrueMNIS} shows the denoising process for both corruption models. The top row denotes the denoising from random vertical and horizontal lines, while the bottom row corresponds to the random sine waves. 

As we can see, the GAN used in estimating the clean images from structured noises is able to generate clean images after training of generator and discriminator. 
In the next section, we use our trained generator for denoising of new images which have not been used during the training.

\subsection{Denoising from Structured Corruption -- Testing}
\begin{figure*}
\begin{tabular}{ccccccccccccccccc}
\hspace{-.4cm}
\begin{tabular}{cccccc} 
\multirow{7}{*}{\tiny{\textbf{\tiny{Corrupted digits}}}}&
&&\includegraphics [width=.08\linewidth]{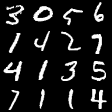}&&\\
\multirow{7}{*}{\tiny{\textbf{\tiny{denoising-GAN}}}} &
\includegraphics [width=.08\linewidth]{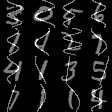}&
\includegraphics [width=.08\linewidth]{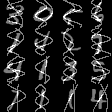}&
\includegraphics [width=.08\linewidth]{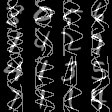}&
\includegraphics [width=.08\linewidth]{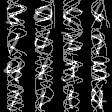}&
\includegraphics [width=.08\linewidth]{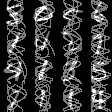}\\ &
%\multirow{7}{*}{\tiny{Vanila GAN}} &
\includegraphics [width=.08\linewidth]{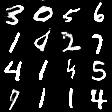}& 
\includegraphics [width=.08\linewidth]{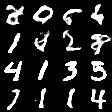}& 
\includegraphics [width=.08\linewidth]{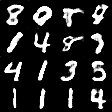}& 
\includegraphics [width=.08\linewidth]{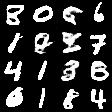}&
\includegraphics [width=.08\linewidth]{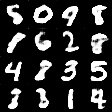}\\
\end{tabular}& & & & & & & & & & & & & & & &
\hspace{-.7cm}
\begin{tabular}{cccccc}
&&&\includegraphics [width=.08\linewidth]{./All_talks/test_True_wave1.png}&&\\&
\includegraphics [width=.08\linewidth]{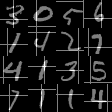}&
\includegraphics [width=.08\linewidth]{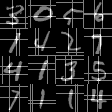}&
\includegraphics [width=.08\linewidth]{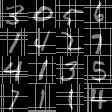}&
\includegraphics [width=.08\linewidth]{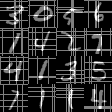}&
\includegraphics [width=.08\linewidth]{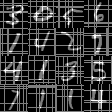}\\&
\includegraphics [width=.08\linewidth]{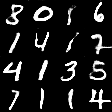}& 
\includegraphics [width=.08\linewidth]{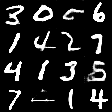}& 
\includegraphics [width=.08\linewidth]{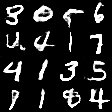}& 
\includegraphics [width=.08\linewidth]{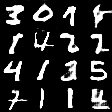}& 
\includegraphics [width=.08\linewidth]{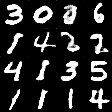}\\
\end{tabular}
\end{tabular}

\begin{tabular}
{ccccccccccccccccccccccccccccccccccccccccccccccccccccccccccccccccccccccccccccccccccccccccccccccccccccccccccc}
&&&&&&&&&&&&&&&&&&&&&&&&&&&&
\begin{tikzpicture}[scale=0.47]
\begin{axis}[
                width=5cm,
                height=4.5cm,
                scale only axis,
                xmin=1, xmax=5,
                xlabel = {Level of Corruption},
                xmajorgrids,
                ymin=0, ymax=1,
                ylabel={Accuracy of Classifier},
                ymajorgrids,
                line width=1.0pt,
                mark size=1.5pt,
                legend style={at={(0.29,.77)},anchor=south west,draw=black,fill=white,align=left}
                ]
\addplot  [color=blue,
               solid, 
               very thick,
               mark=o,
               mark options={solid,scale=1.3},
               ]
               table [x index=0,y index=1]{wa1};
\addlegendentry{Corrupted Images}
\addplot [color=red,
               solid, 
               very thick,
               mark=square,
               mark options={solid,scale=1.3},
               ]
               table [x index=0,y index=2]{wa1};
\addlegendentry{Denoised Images}             
\end{axis}
\end{tikzpicture}& 
%%%%%%%%%%%%%%%%%%%%%%%%%%%%
\begin{tikzpicture}[scale=0.47]
\begin{axis}[
                width=5cm,
                height=4.5cm,
                scale only axis,
                xmin=1, xmax=5,
                xlabel = {Level of Corruption},
                xmajorgrids,
                ymin=0, ymax=.15,
                ylabel={Reconstruction Error (per pixel)},
                ymajorgrids,
                line width=1.0pt,
                mark size=1.5pt,
                legend style={at={(0.34,.77)},anchor=south west,draw=black,fill=white,align=left}
                ]
\addplot  [color=blue,
               solid, 
               very thick,
               mark=o,
               mark options={solid,scale=1.3},
               ]
               table [x index=0,y index=1]{MSE_denoGAN};
\addlegendentry{denoising-GAN}          
\end{axis}
\end{tikzpicture}
& & & & & & & & & & & &&&&&&
\begin{tikzpicture}[scale=0.47]
\begin{axis}[
                width=5cm,
                height=4.5cm,
                scale only axis,
                xmin=1, xmax=5,
                xlabel = {Level of Corruption},
                xmajorgrids,
                ymin=0, ymax=1,
                ylabel={Accuracy of Classifier},
                ymajorgrids,
                line width=1.0pt,
                mark size=1.5pt,
                legend style={at={(0.0,.0)},anchor=south west,draw=black,fill=white,align=left}
                ]
\addplot  [color=blue,
               solid, 
               very thick,
               mark=o,
               mark options={solid,scale=1.3},
               ]
               table [x index=0,y index=1]{hv1};
\addlegendentry{Corrupted Images}
\addplot [color=red,
               solid, 
               very thick,
               mark=square,
               mark options={solid,scale=1.3},
               ]
               table [x index=0,y index=2]{hv1};
\addlegendentry{Denoised Images}             
\end{axis}
\end{tikzpicture}&
%%%%%%%%%%%%%%%%%%%%%%%%%%%%
\begin{tikzpicture}[scale=0.47]
\begin{axis}[
                width=5cm,
                height=4.5cm,
                scale only axis,
                xmin=1, xmax=5,
                xlabel = {Level of Corruption},
                xmajorgrids,
                ymin=0, ymax=.15,
                ylabel={Reconstruction Error (per pixel)},
                ymajorgrids,
                line width=1.0pt,
                mark size=1.5pt,
                legend style={at={(0.34,.77)},anchor=south west,draw=black,fill=white,align=left}
                ]
\addplot  [color=blue,
               solid, 
               very thick,
               mark=o,
               mark options={solid,scale=1.3},
               ]
               table [x index=0,y index=2]{MSE_denoGAN};
\addlegendentry{denoising-GAN}          
\end{axis}
\end{tikzpicture}
& & & & & & & & & & &  & & & & & & & & & & & & & & & & & & & & & & & & & & & & & & &&&&&&&&&&&&&&&&&
\end{tabular}
\caption{\small{The performance of trained generator for various \emph{level of corruption ($lc$)} in denoising of unseen images. Top row: Ground truth digits together with corrupted digits with random sinusoids, and vertical and horizontal lines with a level of corruption from $1$ to $5$. Bottom row: Classification accuracy of pre-trained MNIST classifier for both corrupted and denoised digits along with the reconstruction error per pixel.}}
\label{TESTdenoise}
\end{figure*}

Now we test our framework with test corrupted images for both models of corruption introduced above. For reconstructing of the clean images, we solve the optimization problem in~(\ref{denoisetest}) to obtain solution $\widehat{z}$. Then we find the reconstructed clean images by evaluating $G_{\widehat{\theta}_g}(\widehat{z})$. In Figures~\ref{TESTdenoise}, we have used the different level of corruptions for both of the corruption models. In the top right, we vary the level of corruption from $1$ to $5$ with random sines. The result of $G_{\widehat{\theta}_g}(\widehat{z})$ has been shown below of each level of corruption. In the top left, we have a similar experiment with various level of vertical and horizontal corruptions. Also, we show the denoised images in the below of the corrupted ones. As we can see, even with heavily corrupted images (level corruption equals to $5$), GAN is able to remove the corruption from unseen images and reconstruct the clean images.  
In the bottom row of Figure~\ref{TESTdenoise}, we evaluate the quality of reconstructed images compared to the corrupted ones through a classification task. That is, we use a pre-trained model for MNIST classifier which has test accuracy around $\%98$\footnote{The architecture comprises of two initial convolutional layers along with max-pooling followed by a fully connected layer and a dropout on top of it. Relu is used for all the activation functions and a soft-max function is used in the last layer.}. We feed the MNIST classifier with both denoised (output of denoising-GAN) and corrupted images with a different level of corruptions. For the ground truth labels, we use the labels corresponding to the images before corruption. One interesting point is that when the level of corruption increases the denoised digits are sometimes tweaked compared to the ground truth. That is, in pixel-level, they might not close to the ground truth; however, semantically they are the same. We have also plot the reconstruction error per pixel (normalized by 16 images) for various lc's.

\subsection{Demixing of Structured Signals -- Training} 
\label{mnistTrain}
In this section, we present the results of our experiments for demixing of the structured components. To do this, we use the proposed architecture in Figure~\ref{arcdonisedemix}(b). We first present our experiment with MNIST dataset, and then we show the similar set of experiments with Fashion-MNIST dataset (F-MNIST)~\cite{xiao2017}. This dataset includes $60000$ training $28\times 28$ gray-scale images with $10$ labels. The different labels denote objects, including T-shirt/top, Trouser, Pullover, Dress, Coat, Sandal, Shirt, Sneaker, Bag, and Ankle boot. 

\subsubsection{Experiments on MNIST Dataset}
\begin{figure}[h]
\centering
\begin{tabular}{ccccc}
\multirow{4}{*}{\includegraphics [width=0.15\linewidth]{corrup_64_sin1.png}}&
\includegraphics [width=.17\linewidth]{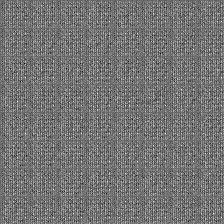}&
\includegraphics [width=.17\linewidth]{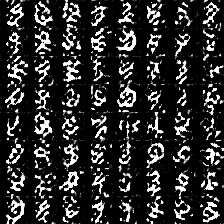}&
\includegraphics [width=.17\linewidth]{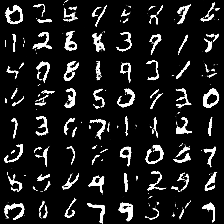} & 
\includegraphics [width=.17\linewidth]{./All_talks/gen1_sin/x_sample_55.png} \\&
\includegraphics [width=.17\linewidth]{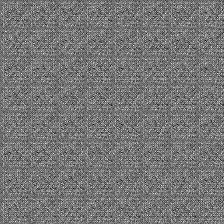}&
\includegraphics [width=.17\linewidth]{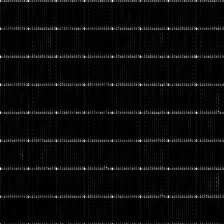}&
\includegraphics [width=.17\linewidth]{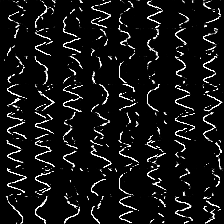}& 
\includegraphics [width=.17\linewidth]{./All_talks/gen2_sin/x_sample_55.png}\\
\\
\end{tabular}
\hspace{1.2cm}
\begin{tabular}{ccccc}
\multirow{4}{*}{\includegraphics [width=0.15\linewidth]{corrup_64_hv.png}}&
\includegraphics [width=.17\linewidth]{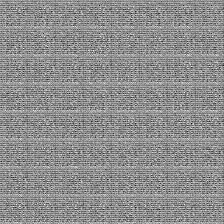}&
\includegraphics [width=.17\linewidth]{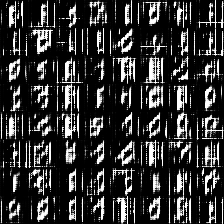}&
\includegraphics [width=.17\linewidth]{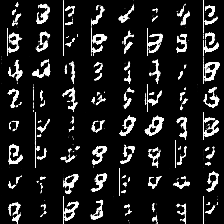}& 
\includegraphics [width=.17\linewidth]{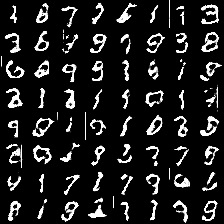}\\&
\includegraphics [width=.17\linewidth]{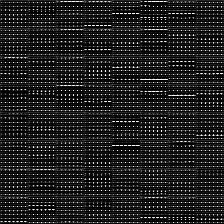}&
\includegraphics [width=.17\linewidth]{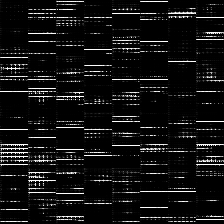}&
\includegraphics [width=.17\linewidth]{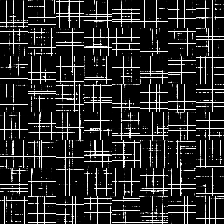} &
\includegraphics [width=.17\linewidth]{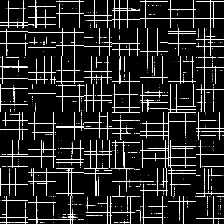} \\
\tiny{Mixed images}&\tiny{$1^{st}$ epoch} & \tiny{$2^{nd}$ epoch} & \tiny{$5^{th}$ epoch} & \tiny{$64^{th}$ epoch}
\end{tabular}
\caption{\small{Evolution of output samples by two generators for fixed $z_1$ and $z_2$. The top panel shows the evolution of the two generators in different epochs where the mixed images comprise of digits and sinusoidal. The first generator is learning the distribution of MNIST digits, while the second one is learning the random sinusoidal waves. The bottom panel shows the same experiment with random horizontal and vertical lines as the second components in the mixed images.}}
\label{DemixingTrain}
\end{figure}

We start the experiments with considering four sets of constituent components. In the first two, similar to the denoising case, we use both random sinusoidal waves and random vertical and horizontal lines for the second constituent component. The difference here is that we are interested in generating of the samples from the second component as well. In Figure~\ref{DemixingTrain}, we show the training evolution of two fixed random vectors, $z_1$ and $z_2$ in $\R^{100}$ in the output of two generators. In the top panel, we have added one random sinusoidal waves to the clean images. As we can see, our proposed GAN architecture can learn two distributions and generate samples from each of them. In the bottom panel, we repeat the same experiment with random vertical and horizontal lines as the second component (two random vertical and two random horizontal lines are added to the clean images). While there is some notion of mode collapse, still two generators can produce the samples from the distribution of the constituent components.

\begin{figure}[h]
\centering
\begin{tabular}{ccccc}
\multirow{2}{*}[1em]{\includegraphics [width=0.15\linewidth]{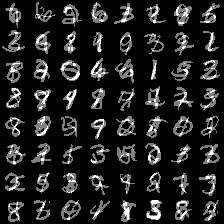}}&
\includegraphics [width=.17\linewidth]{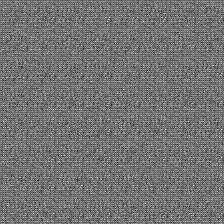}&
\includegraphics [width=.17\linewidth]{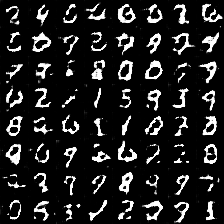}&
\includegraphics [width=.17\linewidth]{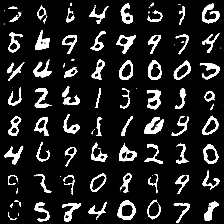}& 
\includegraphics [width=.17\linewidth]{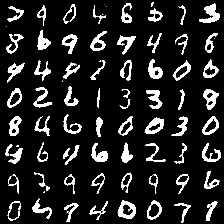}\\&
\includegraphics [width=.17\linewidth]{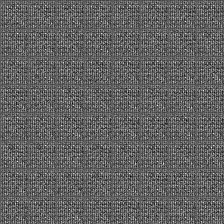}&
\includegraphics [width=.17\linewidth]{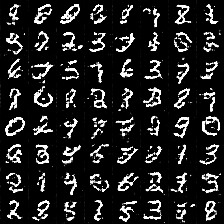}&
\includegraphics [width=.17\linewidth]{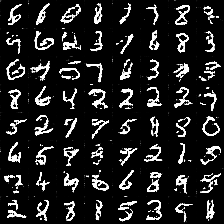} &
\includegraphics [width=.17\linewidth]{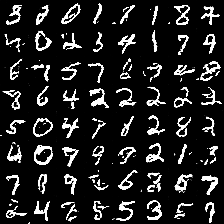} \\
\\
\end{tabular}
\begin{tabular}{ccccc}
\multirow{2}{*}[1em]{\includegraphics [width=0.15\linewidth]{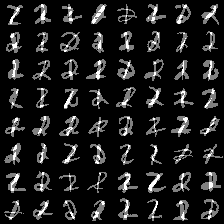}}&
\includegraphics [width=.17\linewidth]{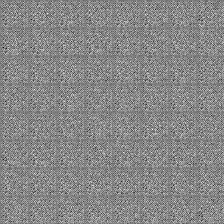}&
\includegraphics [width=.17\linewidth]{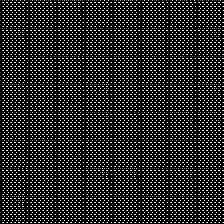}&
\includegraphics [width=.17\linewidth]{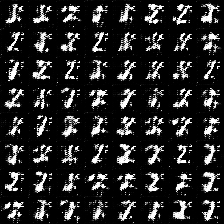}& 
\includegraphics [width=.17\linewidth]{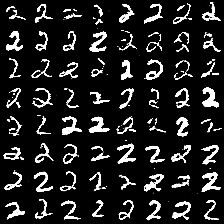}\\&
\includegraphics [width=.17\linewidth]{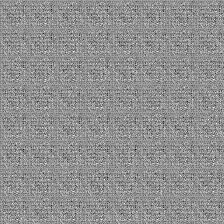}&
\includegraphics [width=.17\linewidth]{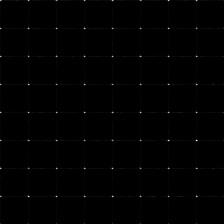}&
\includegraphics [width=.17\linewidth]{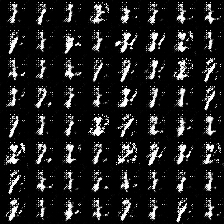} &
\includegraphics [width=.17\linewidth]{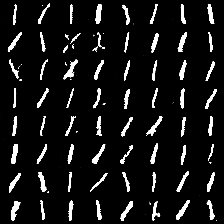} \\
\tiny{Mixed images}&\tiny{$1^{st}$ epoch} & \tiny{$6^{th}$ epoch} & \tiny{$15^{th}$ epoch} & \tiny{$32^{th}$ epoch}
\end{tabular}
\caption{\small{Evolution of output samples by two generators for fixed $z_1$ and $z_2$. The top panel shows that each generator is learning the distribution of one digit out of all $10$ possible digits. The mixed images comprise two arbitrary digits between $0$ to $9$. The bottom panel panel shows a similar experiment where the mixed images comprise only digits $1$ and $2$.}}
\label{digits_DemixingTraining}
\end{figure}

In the second scenario, our mixed images comprise of two MNIST digits from $0$ to $9$. In this case, we are interested in learning the distribution from which each of the digits is drawn. The Top panel in Figure~\ref{digits_DemixingTraining} shows the evolution of
two fixed random vectors, $z_1$ and $z_2$. As we can see, after $32$ epoch, the output of the generators would be the samples of 
MNIST digits. Finally, in the last scenario, we generate the mixed images as the superposition of digits $1$ and $2$. In the training set of MNIST dataset, there are around $6000$ samples from each digits of $1$ and $2$. We have used these digits to form the set of superposition images. The bottom panel of Figure~\ref{digits_DemixingTraining} shows the output of two generators, which can learn the distribution of the two digits. The interesting point is that these experiments show that each GAN can learn the existing digit variety in MNIST training dataset, and we typically do not see mode collapse, which is a major problem in the training of GANs~\cite{goodfellow2016nips}.

\subsubsection{Experiments on F-MNIST Dataset}
\label{fmnistTrain}

\begin{figure}[h]
\centering
\begin{tabular}{ccccc}
\multirow{1}{*}[2em]{\includegraphics [width=0.15\linewidth]{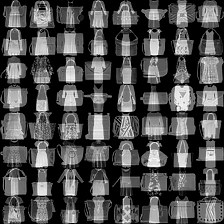}}&
\includegraphics [width=0.17\linewidth]{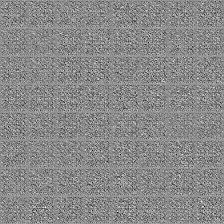}&
\includegraphics [width=0.17\linewidth]{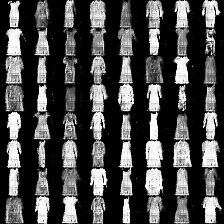}&
\includegraphics [width=0.17\linewidth]{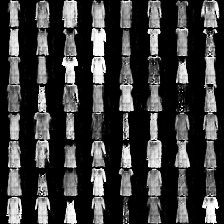}& 
\includegraphics [width=0.17\linewidth]{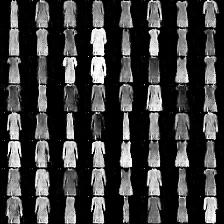}\\&
\includegraphics [width=0.17\linewidth]{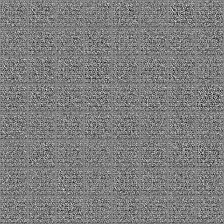}&
\includegraphics [width=0.17\linewidth]{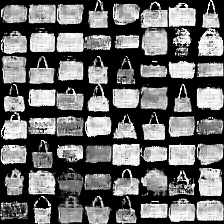}&
\includegraphics [width=0.17\linewidth]{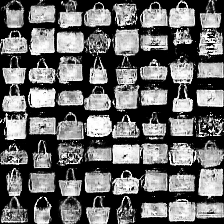} &
\includegraphics [width=0.17\linewidth]{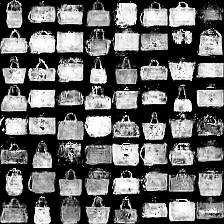} \\
\tiny{Mixed images}&\tiny{$1^{st}$ epoch} & \tiny{$6^{th}$ epoch} & \tiny{$15^{th}$ epoch} & \tiny{$21^{th}$ epoch}
\end{tabular}
\caption{\small{Evolution of output samples by two generators for fixed $z_1$ and $z_2$. The mixed images comprise only two objects, dress, and bag in training F-MNIST dataset. One generator produces the samples from dress distribution, while the other one outputs the samples from the bag distribution.}}
\label{FMNIST_DemixingTraining_two}
\end{figure}

In this section, we illustrate the performance of the proposed demixing-GAN for another F-MNIST dataset. This dataset includes $60000$ training $28\times 28$ gray-scale images with $10$ labels. The different labels denote objects, including T-shirt/top, Trouser, Pullover, Dress, Coat, Sandal, Shirt, Sneaker, Bag, and Ankle boot. Similar to the experiment with MNIST dataset being illustrated in Figure~\ref{digits_DemixingTraining}, we train the demixing-GAN where we have used InfoGAN~\cite{chen2016infogan} architecture for the generators. The architecture of the generators in InfoGAN is very similar to the DCGAN discussed above with the same initialization procedure. The dimension of input noise to the generators is set to $62$. We have also used the same discriminator in DCGAN. Figure~\ref{FMNIST_DemixingTraining_two} shows the output of two generators, which can learn the distribution of dress and bag images during $21$ epochs.

\begin{table*}[htb]
\centering
\begin{tabular}{|c|c|c|c|c|}\hline
&MSE ($1^{st}$ Part) & MSE ($2^{nd}$ Part) & PSNR ($1^{st}$ Part) & PSNR ($2^{nd}$ Part) \\\hline
First Row & 0.04715 & 0.03444 & 13.26476 & 14.62877 \\\hline
Second Row & 0.04430 & 0.03967 & 13.53605 & 14.01344\\\hline
Third Row & 0.05658 & 0.05120 & 12.47313 & 12.90715\\\hline
Forth Row & 0.08948 & 0.10203 & 10.48249 & 9.91264\\\hline
\end{tabular}
\caption{\small{Numerical Evaluation of the results in Figure~\ref{DemxingTest} according to the \emph{Mean Square Error (MSE)} and \emph{Peak Signal-to-Noise ratio (PSNR)} criteria between the corresponding components in the superposition model.}}
\label{msemnist}
\end{table*}

\begin{table*}[t]
\centering
\begin{tabular}{|c|c|c|c|c|}\hline
&MSE ($1^{st}$ Part) & MSE ($2^{nd}$ Part) & PSNR ($1^{st}$ Part) & PSNR ($2^{nd}$ Part) \\\hline
First Row & 0.16859 & 0.12596 & 7.73173 & 8.99763 \\\hline
Second Row & 0.05292 & 0.03304 & 12.76368 & 14.80992\\\hline
Third Row & 0.13498 & 0.11758 & 8.69732 & 9.29655\\\hline
Fourth Row & 0.12959 & 0.08727 & 8.87432 & 10.59132\\\hline
Fifth Row &  0.18250 & 0.12221 & 7.38733 & 9.12906\\\hline
\end{tabular}
\caption{\small{Numerical Evaluation of the results in Figure~\ref{FMNIST_DemixingTest_two} according to the \emph{Mean Square Error (MSE)} and \emph{Peak Signal-to-Noise ratio (PSNR)} criteria between the corresponding components in the superposition model.}}
\label{meFmnist}
\end{table*}

\subsection{Demixing of Structured Signals -- Testing} 
\label{TestMnist}

\begin{figure}[h]
\centering
\begin{tabular}{cccccc}
 \multirow{1}{*}[2em]{\tiny{\textbf{demixing-GAN  }}}&
  \includegraphics [width=0.13\linewidth]{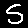}& 
  \includegraphics [width=0.13\linewidth]{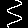}& 
  \includegraphics [width=0.13\linewidth]{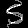}&
  \includegraphics [width=0.13\linewidth]{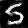}&
  \includegraphics [width=0.13\linewidth]{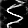}\\
 \multirow{1}{*}[2em]{\tiny{\textbf{demixing-GAN  }}}&
  \includegraphics [width=.13\linewidth]{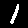}& 
  \includegraphics [width=.13\linewidth]{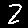}& 
  \includegraphics [width=.13\linewidth]{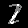}&
  \includegraphics [width=.13\linewidth]{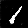}&
  \includegraphics [width=.13\linewidth]{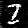}\\
 \multirow{1}{*}[2em]{\tiny{\textbf{ICA  }}}&
  \includegraphics [width=.13\linewidth]{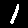}& 
  \includegraphics [width=.13\linewidth]{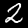}& 
  \includegraphics [width=.13\linewidth]{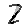}&
  \includegraphics [width=.13\linewidth]{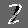}&
  \includegraphics [width=.13\linewidth]{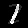}\\
  &\tiny{$1^{st}$ Part}&\tiny{$2^{nd}$ Part}&\tiny{Mixed Image}&\tiny{Est. $1^{st}$ Part}&\tiny{ Est. $2^{nd}$ Part}
\end{tabular}
\caption{\small{ The performance of trained generators for demixing of two constituent components. The first two columns are the ground-truth components. The third column is the ground-truth mixed image and the last two columns denote the recovered components. The first row uses the same generator trained for only one digit (drawn from MNIST test dataset) and a random sinusoidal. The second row uses the generator trained only for digits $1$ and $2$. The last row shows the result of demixing with ICA method.}}
\label{DemxingTest}
\end{figure}

Similar to the test part of denoising, in this section, we test the performance of two trained generators in a demixing scenario for the mixed images, which have not been seen in the training time. Figure~\ref{DemxingTest} shows our third experiment in which we have illustrated the demixing on three different input mixed images. Here, we have compared the performance of demixing-GAN with \emph{Independent component analysis (ICA)} method~\cite{hoyer1999independent}. In the top and middle rows of Figure~\ref{DemxingTest}, we consider the mixed images generated by adding a digit (drawn from  MNIST test dataset) and a random sinusoidal. Then the goal is to separate (demix) these two from the given superimposed image. To do this, we use GAN trained  for learning the distribution of digits and sinusoidal waves (the top panel of Figure~\ref{DemixingTrain}) and solve the optimization problem in (\ref{demixtest}) through an alternative minimization fashion. As a result, we obtain $\widehat{z_1}$ and $\widehat{z_2}$. The corresponding constituent components is then obtained by evaluating $G_{\widehat{\theta}_{g_1}}(\widehat{z_1})$ and $G_{\widehat{\theta}_{g_2}}(\widehat{z_2}$). In Figure~\ref{DemxingTest}, the first two columns denote the ground-truth of the constituent components. The middle one is the mixed ground-truth, and the last two show the recovered components using demixing-GAN and ICA. In the last row, digits $1$ and $2$ drawn from the MNIST test dataset are added to each other and we apply the GAN trained for learning the distribution of digits $1$ and $2$ (bottom panel in Figure~\ref{digits_DemixingTraining}). As we can see, our proposed GAN can separate two digits; however, ICA method fails in demixing of two components. In addition, Table~\ref{msemnist} has compared numerically the quality of recovered components with the corresponding ground-truth ones through \emph{mean square error (MSE)} and \emph{Peak Signal-to-Noise Ratio (PSNR)} criteria. 

\subsubsection{Demixing of F-MNIST -- Testing}

\begin{figure}[h]
\centering
\begin{tabular}{cccccc}
 \multirow{1}{*}[1.5em]{\tiny{\textbf{demixing-GAN}}}&
\includegraphics [width=0.12\linewidth]{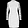}&
\includegraphics [width=0.12\linewidth]{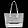}&
\includegraphics [width=0.12\linewidth]{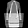}& 
\includegraphics [width=0.12\linewidth]{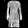}&
\includegraphics [width=0.12\linewidth]{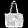}\\
 \multirow{1}{*}[1.5em]{\tiny{\textbf{demixing-GAN}}}&
\includegraphics [width=0.12\linewidth]{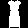}&
\includegraphics [width=0.12\linewidth]{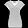}&
\includegraphics [width=0.12\linewidth]{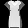}& 
\includegraphics [width=0.12\linewidth]{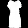}&
\includegraphics [width=0.12\linewidth]{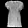}\\
 \multirow{2}{*}[1.5em]{\tiny{\textbf{demixing-GAN}}}&
\includegraphics [width=0.12\linewidth]{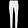}&
\includegraphics [width=0.12\linewidth]{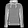}&
\includegraphics [width=0.12\linewidth]{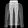}& 
\includegraphics [width=0.12\linewidth]{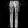}&
\includegraphics [width=0.12\linewidth]{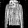}\\
 \multirow{1}{*}[1.5em]{\tiny{\textbf{demixing-GAN}}}&
\includegraphics [width=0.12\linewidth]{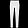}&
\includegraphics [width=0.12\linewidth]{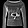}&
\includegraphics [width=0.12\linewidth]{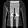}& 
\includegraphics [width=0.12\linewidth]{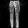}&
\includegraphics [width=0.12\linewidth]{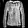}\\
 \multirow{1}{*}[1.5em]{\tiny{\textbf{ICA}}}&
\includegraphics [width=0.12\linewidth]{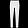}& 
\includegraphics [width=0.12\linewidth]{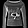}& 
\includegraphics [width=0.12\linewidth]{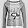}&
\includegraphics [width=0.12\linewidth]{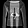}&
\includegraphics [width=0.12\linewidth]{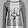}\\
&\tiny{$1^{st}$ Part}&\tiny{$2^{nd}$ Part} & \tiny{Mixed image} & \tiny{Est. $1^{st}$ Part} & \tiny{Est. $2^{nd}$ Part}
\end{tabular}
\caption{\small{The performance of trained generators for demixing of two constituent components. The first two columns are the ground-truth components. The third column is the ground-truth mixed image and the last two columns denote the recovered components. The first row uses the generator trained for only two objects for $20$ epochs. The second row uses the generator trained for all $10$ objects for $20$ epochs. The third and fourth rows use the same generator trained for only two objects for $30$ epochs. The last row shows the result of demixing with ICA method.}}
\label{FMNIST_DemixingTest_two}
\end{figure}

In this section, we evaluate the performance of trained demixing-GAN on the F-MNIST dataset. In Figure~\ref{FMNIST_DemixingTest_two}, we have illustrated an experiment similar to the setup in section~\ref{TestMnist}. The first two columns in Figure~\ref{FMNIST_DemixingTest_two} denote two objects from F-MNIST test dataset as the ground-truth components. The third column is the ground-truth mixed image, and the last two columns show the recovered constituent components. The first row uses the generator trained for only two objects for $20$ epochs. The second row uses the generator trained for all $10$ objects for $20$ epochs. The third and fourth rows use the same generator trained for only two objects for $30$ epochs. The last row shows the result of demixing with ICA method. We have implemented ICA using Scikit-learn module~\cite{scikit-learn}. As we can see, ICA fails to separate the components (images of F-MNIST) from each other, while the proposed demixing-GAN can separate the mixed images from each other. However, the estimated image components are not exactly matched to the ground-truth ones (first two columns). This has been shown through numerical evaluation according to MSE and PSNR in Table~\ref{meFmnist}. 

Finally, as an attempt to understand the condition under which the demixing-GAN is failed, we empirically investigated the role of two aspects pf demixing-GAN. First, it seems that the hidden space ($z$-space) of the generators for characterizing the distribution of the constituent components play an essential role in the success/failure of the demixing-GAN. Second the incoherent hidden structures in the components in the generator space. We investigate these observations through some numerical experiments in section~\ref{FailOfDemixing} in the appendix. 

\section{Conclusion}
In this paper, we considered the GANs framework for learning the structure of the constituent components in a superposition observation model. We empirically showed that it is possible to implicitly learn the underlying distribution of each component and use them in the downstream task such as denoising, demixing, and sparse recovery. We also investigate the conditions under which the proposed demixing framework fails through extensive experimental simulations and some theoretical insights.  

%% file: append.tex
\section{Appendix}
\label{append}

\subsection{Some Theoretical Intuitions about the demixing-GAN}
Recall that the superposition model is given by $Y= X + N$, and $\D_y$, $\D_x$ and $\D_n$ denote the distribution of $Y, X$, and $N$, respectively. Let $G_1(z_1)\overset{\Delta}{=}G_{{\theta}_{g_1}}({z_1}) \sim\D_{g_1}$ and $G_2(z_2)\overset{\Delta}{=}G_{{\theta}_{g_2}}({z_2})\sim\D_{g_2}$. Also assume that $(z_1, z_2)\sim\D_{z_1,z_2}$ denotes the joint distribution of the hidden random vectors with marginal probability as $\D_{z_i}$ for $i=1,2$. We note that in demxing setting there are not samples from the component $N$ as opposed to the denoising scenario. Now we have the following mini-max loss as \eqref{GANprob}:
\begin{align}
\label{Theorydemixing-GAN}
\min_{G_1,G_2}\max_{D}\mathcal{L}(G_1, G_2, D) = \E_{u\sim\D_y}log(D(u)) + \E_{(z_1, z_2)\sim\D_{z_1,z_2}}log(1-D(G_1(z_1) + G(z_2))).
\end{align}
Following the standard GAN framework, for the fixed $G_1$ and $G_2$, we have:
\begin{align*}
\mathcal{L}(G_1, G_2, D) = \int_u(\D_y(u)log(D(u)) + \D_G(u)log(1-D(u)))du, 
\end{align*}
where $\D_G = \D_{g_1}*\D_{g_2}$. Hence, the optimal discriminator is given by $D^* = \frac{\D_x*\D_n}{\D_x*\D_n + \D_{g_1}*\D_{g_2}}$ since $Y = X+N$ and the fact that $\D_y$ and $\D_g$ are the pdf and defined in $[0,1]$. This means that the global optimal of problem~(\ref{Theorydemixing-GAN}) is achieved iff $\D_x*\D_n = \D_{g_1}*\D_{g_2}$ ( $*$ denotes the convolution operator). However, this condition is generally an ill-posed equation. That is, in general, $\D_x\neq \D_{g_1}$ and $\D_n\neq\D_{g_2}$. In the best case, we can have hope to uniquely determine the distributions up to a permutation (similar thing is also true for the ICA method). This is the point actually we need some notion of incoherence between two constituent structures, $\X$, and $\N$. So, the question is this under what incoherent condition, we can have a well-conditioned equation? According to our previous discussion in the denoising case, even if we somehow figure out the right incoherent condition, $\D_x$ is uniquely determined by $\D_n$ if the Fourier transform of $\D_n$ is non-zero. While we currently do not have a right answer for the above question, we conjecture that in addition to the incoherence issue in the signal domain, the hidden space ($z$-space) in both generators play an important role to make the demixing problem possible. We investigate this idea and the other things empirically at the end of the appendix.      

\subsection{Detail of Experiments}

Here, we give some more details about the initialization of the generators we have used for the MNIST and F-MNIST experiments in sections~\ref{mnistTrain} and ~\ref{TestMnist}. As we mentioned earlier, we have used the same architecture based on the DCGAN for both of the generators used in the demixing-GAN for MNIST dataset. In particular, the generators include three layers: the first two layers are fully connected layers with Relu activation function. We have not used any batch-normalization (as opposed to the original DCGAN). Also, the weights in these fully connected layers are initialized according to the random normal distribution with standard deviation equals to $0.02$. We have also used zero initialization for the biases. The third layer includes a transposed convolution layer with filters size of which are set to $5$ and stride to $2$. We have initialized these filters according to the random normal distribution with standard deviation equals to $0.02$. Relu nonlinearity has also been used after the transposed convolution operation. We have not used any max-pooling in this architecture. Similarly, the discriminator comprises two convolution layers with leaky-rely activation function followed by a fully connected layer without any max-pooling. The weights in these convolution layers have been initialized based on the truncated normal distribution with standard deviation equals to $0.02$. The fully connected layer is initialized as before. Finally, we fed two generators with $i.i.d$ random noise vector with entries uniformly drawn from $[-1,1]$. For the F-MNIST experiments, we have used a similar architecture to the InfoGAN generator~\cite{chen2016infogan}. The architecture of the generators is similar to the previous case except that the transposed convolution in the third layer has $128$ filters with size $4\times4$ and stride 2, while the previous DCGAN has $64$ filters with size $5\times5$ and stride 2. The discriminators are the same. For the ICA experiment, we simulate the mixing process simalr to~\cite{anirudh2018unsupervised} using mixing matrix with entries drawn from a truncated random normal distribution, i.e., $W_{ij}\sim\N(-0.5, 0.5)$ for $i,j=1,2$, allowing for negative weights. Hence, the final mixed observation is given by $Y=XW^T$ where $X$ denotes the constituent component matrix (source matrix) with 2 columns (number of components) and with number of rows equal to the size of input mixed image ($28\times28$).   

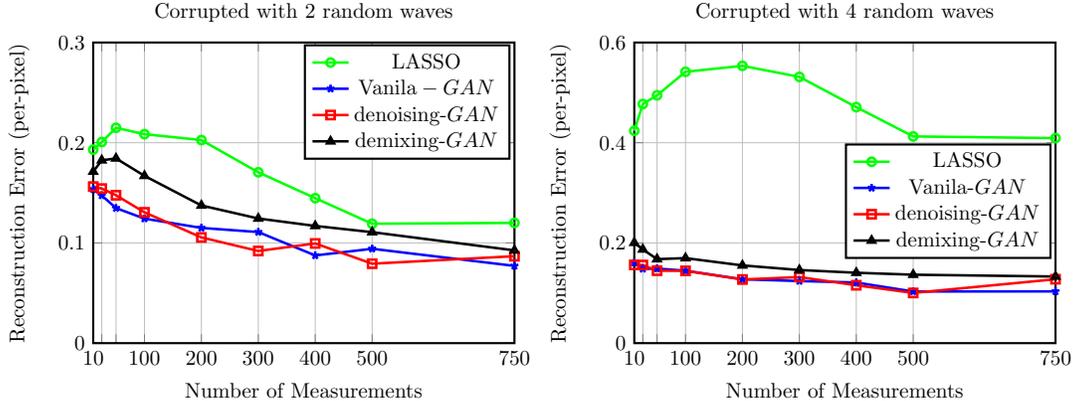
\begin{figure*}[t]
	\centering
	\begin{tabular}{cc}
\begin{tikzpicture}[scale=0.8]
\begin{axis}[
width=7cm,
height=5cm,
scale only axis,
xmin=10, xmax=750,
xlabel = {Number of Measurements },
xtick={10,25,50, 100, 200, 300, 400, 500, 750},
xticklabels={10, , , 100, 200, 300, 400, 500, 750},
xmajorgrids,
ymin=0, ymax= 0.3,
ylabel={Reconstruction Error (per-pixel)},
title={Corrupted with $2$ random waves},
ymajorgrids,
line width=1.0pt,
mark size=1.5pt,
legend style={at={(.50,.61)},anchor=south west,draw=black,fill=white,align=left}
]
\addlegendentry{LASSO}
\addplot  [color=green,
solid, 
very thick,
mark=o,
mark options={solid,scale=1.3},
]
table [x index=0,y index=1]{GANS_CS};
\addlegendentry{$\textrm{Vanila}-GAN$}
\addplot  [color=blue,
solid, 
very thick,
mark=star,
mark options={solid,scale=1.3},
]
table [x index=0,y index=2]{GANS_CS};
\addplot [color=red,
solid, 
very thick,
mark=square,
mark options={solid,scale=1.3},
]
table [x index=0,y index=3]{GANS_CS};
\addlegendentry{$\textrm{denoising-}GAN$}       
\addplot [color=black,
solid, 
very thick,
mark=triangle,
mark options={solid,scale=1.3},
]
table [x index=0,y index=4]{GANS_CS};
\addlegendentry{$\textrm{demixing-}GAN$}     
\end{axis}
\end{tikzpicture}&
\begin{tikzpicture}[scale=0.8]
\begin{axis}[
width=7cm,
height=5cm,
scale only axis,
xmin=10, xmax=750,
xlabel = {Number of Measurements },
xtick={10,25,50, 100, 200, 300, 400, 500, 750},
xticklabels={10, , , 100, 200, 300, 400, 500, 750},
xmajorgrids,
ymin=0, ymax= 0.6,
ylabel={Reconstruction Error (per-pixel)},
title={Corrupted with $4$ random waves},
ymajorgrids,
line width=1.0pt,
mark size=1.5pt,
legend style={at={(.5,.28)},anchor=south west,draw=black,fill=white,align=left}
]
\addlegendentry{LASSO}
\addplot  [color=green,
solid, 
very thick,
mark=o,
mark options={solid,scale=1.3},
]
table [x index=0,y index=1]{GANS_CS2};
\addlegendentry{$\textrm{Vanila-}GAN$}
\addplot  [color=blue,
solid, 
very thick,
mark=star,
mark options={solid,scale=1.3},
]
table [x index=0,y index=2]{GANS_CS2};
\addplot [color=red,
solid, 
very thick,
mark=square,
mark options={solid,scale=1.3},
]
table [x index=0,y index=3]{GANS_CS2};
\addlegendentry{$\textrm{denoising-}GAN$}       
\addplot [color=black,
solid, 
very thick,
mark=triangle,
mark options={solid,scale=1.3},
]
table [x index=0,y index=4]{GANS_CS2};
\addlegendentry{$\textrm{demixing-}GAN$}     
\end{axis}
\end{tikzpicture}
	\end{tabular}
	\caption{\small{Performance of Different GANs in Compressive Sensing Experiments.}}
	\label{CS_GANS}
\end{figure*}

\subsection{Compressive Sensing}
In this section, we present results in the compressed sensing setting. The goal here is to understand the quality of generative models that can be learned under the structured corruption in \emph{denoising}-GAN and \emph{demixing}-GAN architectures.  To this end,  we test the capability of the proposed GANs as the generative model for natural images. The experimental setup is similar to the reconstruction from Gaussian measurement experiment for MNIST data reported in~\cite{bora2017compressed}. In particular, we assume the following observation model by a random sensing matrix $A \in \mathbb{R}^{m \times p}$ with $m < p$ under structured corruption by $N$ as follows:
\begin{align}
Y = A(X + N),
\end{align}
where the entries of $A$ are i.i.d Gaussian with zero mean and variance $\frac{1}{m}$. For a given generator model $G_{\widehat{\theta}}$, we solve the following inference problem: 
\begin{align}
\min_{z} \|  Y - AG_{\widehat{\theta}}(z) \|_2^2  + \lambda \|z\|_2^2,
\label{opt_prob:CS}
\end{align}
where the generator $G_{\widehat{\theta}}$ is taken from various generator networks. For this experiment, we select the signal $X$ randomly from MNIST test dataset, and $N$ from random sinusoidal corruption with various number of waves. We compare generator models learned from clean MNIST images, i.e., generator model obtained from \emph{denoising}-GAN and \emph{demixing}-GAN framework. Both the generator models learned from {denoising}-GAN and {demixing}-GAN were trained under wave corruption model.  For demixing-GAN, we select appropriate GAN by manually looking at the output of generators and choosing the one that gives MNIST like images as output. We also compare our approach to LASSO as MNIST images are naturally sparse. As reported in~\cite{bora2017compressed}. for this experiment the regularization parameter $\lambda$ was set as $0.1$ as it gives the best performance on validation set. For solving the inference problem in \eqref{opt_prob:CS},  we have used ADAM optimizer with step size set to $0.01$. Since the problem is non-convex,  we have used $10$ random initialization with $10000$ iterations. We choose the one gives the best measurement error. 

The results of this experiment are presented in Figure \ref{CS_GANS} where we report per-pixel reconstruction error on $25$ images chosen randomly from the test dataset for two different corruption levels. The plot on the left panel is obtained when the signal is corrupted with $2$ random waves, whereas the right plot corresponds to the corruption with $4$ random waves.  In the both figures, we observe that with corruption, the performance of LASSO significantly degrades. The generator from demixing-GAN improves the performance over LASSO. Generator obtained from denoising-GAN performs comparably to the vanilla GAN. This experiment establishes the quality of the generative models learned from GANs as the prior. The comparable performance of denoising GAN with vanilla GAN shows that meaningful prior can be learned even from heavily corrupted samples.

\subsection{Experiments on F-MNIST Dataset}
\label{fmnistTrain}

\begin{figure}[h]
\centering
\begin{tabular}{ccccc}
\multirow{1}{*}[2em]{\includegraphics [width=0.15\linewidth]{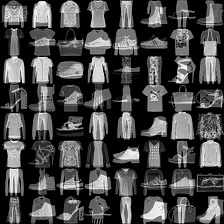}}&
\includegraphics [width=0.17\linewidth]{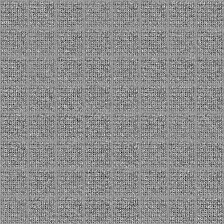}&
\includegraphics [width=0.17\linewidth]{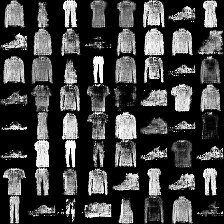}&
\includegraphics [width=0.17\linewidth]{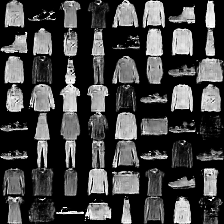}& 
\includegraphics [width=0.17\linewidth]{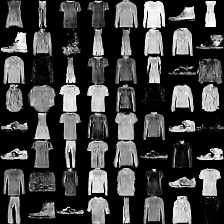}\\&
\includegraphics [width=0.17\linewidth]{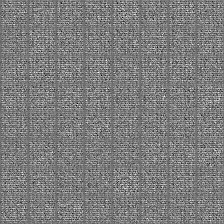}&
\includegraphics [width=0.17\linewidth]{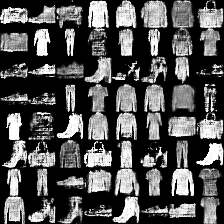}&
\includegraphics [width=0.17\linewidth]{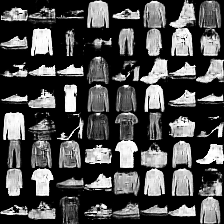} &
\includegraphics [width=0.17\linewidth]{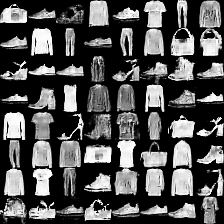} \\
\tiny{Mixed images}&\tiny{$1^{st}$ epoch} & \tiny{$6^{th}$ epoch} & \tiny{$15^{th}$ epoch} & \tiny{$21^{th}$ epoch}
\end{tabular}
\caption{\small{Evolution of output samples by two generators for fixed $z_1$ and $z_2$. The mixed images comprise two arbitrary objects drawn from $10$ objects from training F-MNIST dataset. Each generator outputs the samples from the distribution of all $10$ possible objects.}}
\label{FMNIST_DemixingTraining_all}
\end{figure}

In this section, we present more experiment on F-MNIST dataset. In particular, we construct the mixed images by adding randomly the images in all classes in FMNIST training dataset. As mentioned before, there are 10 different categories in this dataset. Figure~\ref{FMNIST_DemixingTraining_all} shows the evolution of two fixed random vectors, $z_1$ and $z_2$ drawn from $[-1,1]^{62}$. As we can see, after $21$ epoch, the output of the generators would be the samples of F-MNIST objects. We also generate mixed images as the superposition of two objects, dress and bag images. In the training set of the F-MNIST dataset, there are around $6000$ dress and bag images. We have used these images to form the set of superposition images. 

\subsection{Experiments on both MNIST and F-MNIST Dataset}

\begin{figure}[h]
\centering
\begin{tabular}{ccccc}
\multirow{1}{*}[2em]{\includegraphics [width=0.15\linewidth]{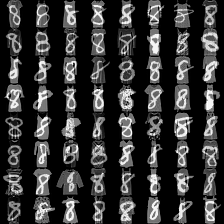}}&
\includegraphics [width=0.17\linewidth]{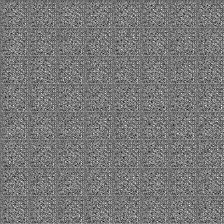}&
\includegraphics [width=0.17\linewidth]{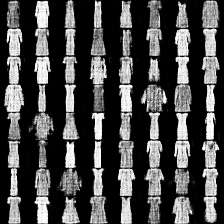}&
\includegraphics [width=0.17\linewidth]{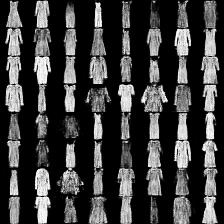}& 
\includegraphics [width=0.17\linewidth]{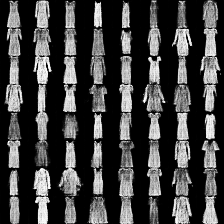}\\&
\includegraphics [width=0.17\linewidth]{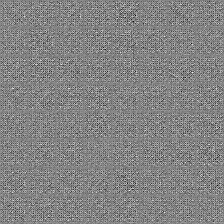}&
\includegraphics [width=0.17\linewidth]{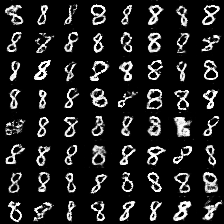}&
\includegraphics [width=0.17\linewidth]{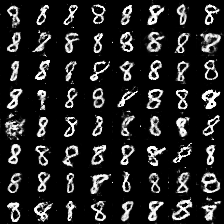} &
\includegraphics [width=0.17\linewidth]{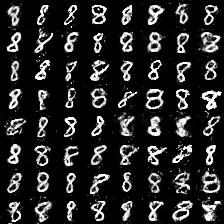} \\
\tiny{Mixed images}&\tiny{$1^{st}$ epoch} & \tiny{$6^{th}$ epoch} & \tiny{$15^{th}$ epoch} & \tiny{$21^{th}$ epoch}
\end{tabular}
\caption{\small{Evolution of output samples by two generators for fixed $z_1$ and $z_2$. The mixed images comprise only two objects, dress, and bag in training F-MNIST dataset. One generator produces the samples from digit $8$ distribution, while the other one outputs the samples from the dress distribution.}}
\label{FMNISTMNIST_DemixingTraining_two}
\end{figure}

In this section, we explore the performance of demixing-GAN when the superposed images comprise the sum of a digit $8$ from MNIST dataset and dress from the F-MNIST dataset. The experiment for this setup has been illustrated in Figure~\ref{FMNISTMNIST_DemixingTraining_two}. Since our goal is to separate dress from the digit $8$, for the first generator, we have used the InfoGAN architecture being used in the experiment in section~\ref{fmnistTrain} and similarly the DCGAN architecture for the second generator as section~\ref{mnistTrain}. As a result, the input noise to the first generator is drawn uniformly from $[-1,1]^{62}$ and uniformly from $[-1,1]^{100}$ for the second generator. Figure~\ref{FMNISTMNIST_DemixingTraining_two} shows the evolution of output samples by two generators for fixed $z_1$ and $z_2$. As we can see, after $21$ epoch, the first generator is able to generate dress samples and the second one outputs samples of digit $8$.  

\subsubsection{Demixing both MNIST and F-MNIST -- Testing}

\begin{figure}
\centering
\begin{tabular}{cccccc}
 \multirow{1}{*}[3em]{\tiny{\textbf{demixing-GAN}}}&
\includegraphics [width=0.12\linewidth]{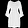}&
\includegraphics [width=0.12\linewidth]{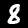}&
\includegraphics [width=0.12\linewidth]{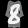}& 
\includegraphics [width=0.12\linewidth]{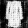}&
\includegraphics [width=0.12\linewidth]{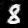}\\
 \multirow{1}{*}[3em]{\tiny{\textbf{ICA}}}&
\includegraphics [width=0.12\linewidth]{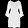}& 
\includegraphics [width=0.12\linewidth]{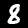}& 
\includegraphics [width=0.12\linewidth]{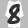}&
\includegraphics [width=0.12\linewidth]{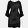}&
\includegraphics [width=0.12\linewidth]{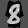}\\
&\tiny{$1^{st}$ Part}&\tiny{$2^{nd}$ Part} & \tiny{Mixed image} & \tiny{Estimated $1^{st}$ Part} & \tiny{Estimated $2^{nd}$ Part}
\end{tabular}
\caption{\small{The performance of trained generators for demixing of two constituent components. The first two columns are the ground-truth components. The third column is the ground-truth mixed image and the last two columns denote the recovered components. The first row uses the generator trained through demixing-GAN. The second row shows the result of demixing with ICA method.}}
\label{MNIST_FMNIST_DemixingTest_two}
\end{figure}

Similar to the previous Testing scenarios, in this section, we evaluate the performance of the demixing-GAN in comparison with ICA for separating a test image which is the superposition of a digit $8$ drawn randomly from MNIST test dataset and dress object drawn randomly from F-MNIST test dataset. Figure~\ref{MNIST_FMNIST_DemixingTest_two} shows the performance of demixng-GAN and ICA method. As we can see, ICA totally fails to demix the two images from each other, whereas the demixing-GAN is able to separate digit $8$ very well and to some extend the dress object from the input superposed image. MSE and PSNR values for the first component using ICA recovery method is given by $0.40364$ and $3.94005$, respectively. Also, MSE and PSNR for the first component using ICA recovery method is given by $0.15866$ and $7.99536$, respectively.  

\subsection{Experiment on Quick-Draw dataset}

\begin{figure}[h]
\centering
\begin{tabular}{ccccc}
\multirow{1}{*}[2em]{\includegraphics [width=0.15\linewidth]{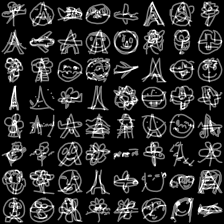}}&
\includegraphics [width=0.17\linewidth]{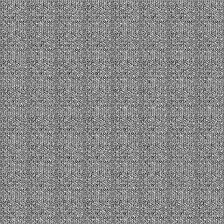}&
\includegraphics [width=0.17\linewidth]{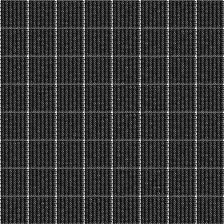}&
\includegraphics [width=0.17\linewidth]{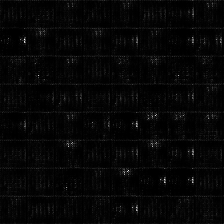}& 
\includegraphics [width=0.17\linewidth]{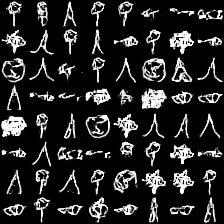}\\&
\includegraphics [width=0.17\linewidth]{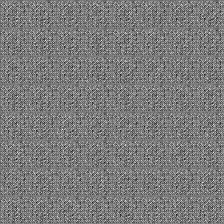}&
\includegraphics [width=0.17\linewidth]{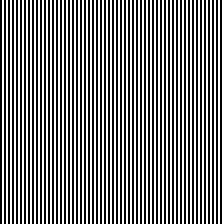}&
\includegraphics [width=0.17\linewidth]{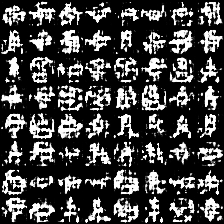} &
\includegraphics [width=0.17\linewidth]{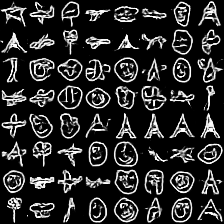} \\
\tiny{Mixed images}&\tiny{$1^{st}$ epoch} & \tiny{$6^{th}$ epoch} & \tiny{$13^{th}$ epoch} & \tiny{$31^{th}$ epoch}
\end{tabular}
\caption{\small{Evolution of output samples by two generators for fixed $z_1$ and $z_2$. The mixed images comprise two arbitrary objects drawn from $10$ objects in training Quick-Draw dataset. Each generator outputs the samples from the distribution of all $5$ objects.}}
\label{QD_DemixingTraining_all}
\end{figure}

In this section, we present our demixing framework in another dataset, Quick-Draw dataset~\cite{QuickDraw} released by Google recently. The Quick Draw Dataset is a collection of 50 million drawings categorized in 345 classes, contributed by players of the game Quick, Draw!~\cite{QuickDrawGame}. For this section, we select 5 classes out of 345 possible categories, including airplane, animal migration, face, flower, and The Eiffel Tower. Also, for each class, we consider 16000 images of size $28\times28$. Figure~\ref{QD_DemixingTraining_all} shows the experiment for all the classes. Similar to MNIST and F-MNIST, we construct the training set by adding all the images randomly from 5 categories. To run this experiment, we feed each generator with random vectors $z_1$ and $z_2$ drawn uniformly from $[-1,1]^{64}$. As illustrated in the figure~\ref{QD_DemixingTraining_all}, after 31 epochs two generators can generate samples from the distribution of all five classes. From this experiment, it seems that  Quick-Draw dataset is more complicated than MNIST or perhaps F-MNIST as it takes more time for demixing-GAN to be able to output resemble samples to the original Quick-Draw objects. 

\begin{figure}[h]
\centering
\begin{tabular}{ccccc}
\multirow{1}{*}[2em]{\includegraphics [width=0.15\linewidth]{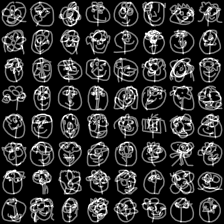}}&
\includegraphics [width=0.17\linewidth]{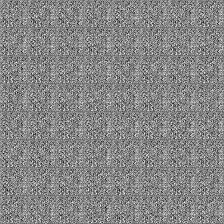}&
\includegraphics [width=0.17\linewidth]{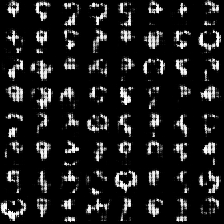}&
\includegraphics [width=0.17\linewidth]{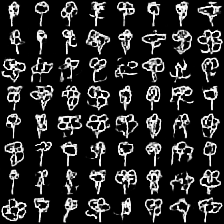}& 
\includegraphics [width=0.17\linewidth]{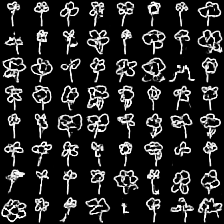}\\&
\includegraphics [width=0.17\linewidth]{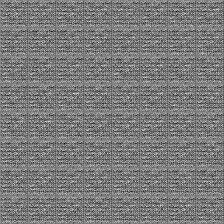}&
\includegraphics [width=0.17\linewidth]{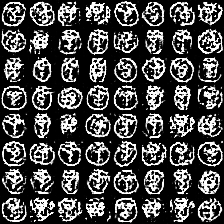}&
\includegraphics [width=0.17\linewidth]{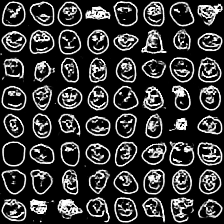} &
\includegraphics [width=0.17\linewidth]{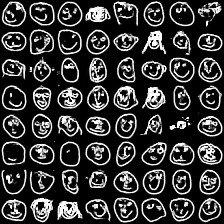} \\
\tiny{Mixed images}&\tiny{$1^{st}$ epoch} & \tiny{$6^{th}$ epoch} & \tiny{$13^{th}$ epoch} & \tiny{$31^{th}$ epoch}
\end{tabular}
\caption{\small{Evolution of output samples by two generators for fixed $z_1$ and $z_2$. The mixed images comprise only two objects, face, and flower in training Quick-Draw dataset. One generator produces the samples from dress distribution, while the other one outputs the samples from the bag distribution.}}
\label{QD_DemixingTraining_two}
\end{figure}

Next, we consider only two objects, face, and flower in the Quick Draw Dataset. As a result, the input mixed images are the superposition of different faces and flowers. Figure~\ref{QD_DemixingTraining_two} shows the evolution of the random vectors $z_1$ and $z_2$ (drawn uniformly from $[-1,1]^{64}$). As we can see, after 31 epochs, one generator can produce various kind of faces, while the other one generates different shapes of flowers.

Finally, we consider a more challenging scenario in which the constituent components in the mixed images are just airplane shapes. That is, we randomly select the airplane shapes from $16000$ images in the training set, and add them together to construct the input mixed images. We have been noticed that in the 16000 images of the airplane shapes, in general, there are two structures. One is related to the airplanes having been drawn by the players in the game Quick, Draw more simply and somehow flat (they are mostly similar to an ellipse with or without wings), while the second one consists the more detailed shapes (they have the tail and maybe with different orientation).  

\begin{figure}[h]
\centering
\begin{tabular}{ccccc}
\multirow{1}{*}[2em]{\includegraphics [width=0.15\linewidth]{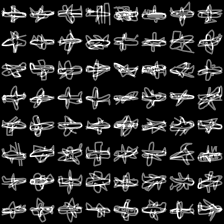}}&
\includegraphics [width=0.17\linewidth]{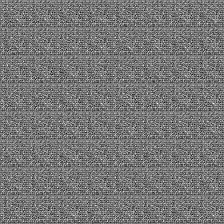}&
\includegraphics [width=0.17\linewidth]{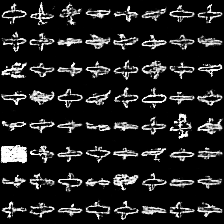}&
\includegraphics [width=0.17\linewidth]{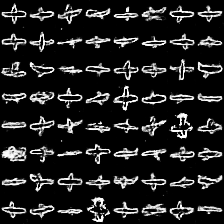}& 
\includegraphics [width=0.17\linewidth]{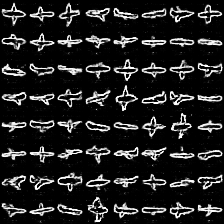}\\&
\includegraphics [width=0.17\linewidth]{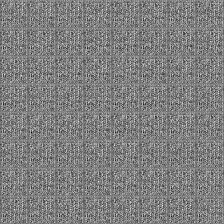}&
\includegraphics [width=0.17\linewidth]{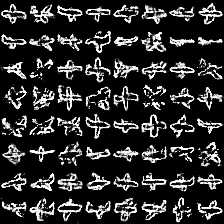}&
\includegraphics [width=0.17\linewidth]{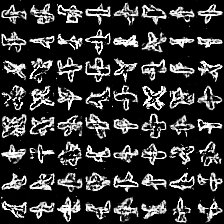} &
\includegraphics [width=0.17\linewidth]{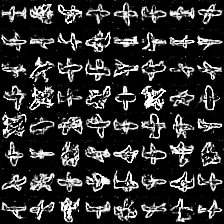} \\
\tiny{Mixed images}&\tiny{$1^{st}$ epoch} & \tiny{$6^{th}$ epoch} & \tiny{$13^{th}$ epoch} & \tiny{$31^{th}$ epoch}
\end{tabular}
\caption{\small{Evolution of output samples by two generators for fixed $z_1$ and $z_2$. The mixed images comprise only airplane object, randomly drawn from the training Quick-Draw dataset. The top generator produces mostly the samples from simpler and flat airplanes, while the bottom one outputs the samples from the more detailed airplane shapes.}}
\label{QD_DemixingTraining_twoair}
\end{figure}

Figure~\ref{QD_DemixingTraining_twoair} depicts the performance of demixing-GAN for this setup. One surprising point is that while both components in the superposition are drawn from one class (e.g., airplane shapes), the demixing-GAN is still able to demix the hidden structure in the airplane distribution. Thus, we think that just having the same distribution for both of the constituent components is not necessarily a barrier for demixing performance. We guess that somehow different features of the shapes drawn from the same distribution makes demixing possible by forcing the incoherence between the components. As we can see, after 31 epochs, both generators can learn two mentioned structures, and regarding two structures, they can cluster the shape of airplanes in two types.   

\subsection{Experiment on SVHN dataset}
\begin{figure}[h]
\centering
\begin{tabular}{ccccc}
\multirow{1}{*}[2em]{\includegraphics [width=0.15\linewidth]{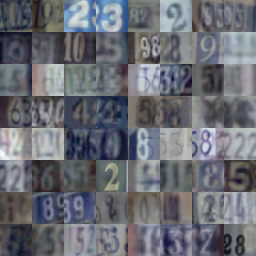}}&
\includegraphics [width=0.17\linewidth]{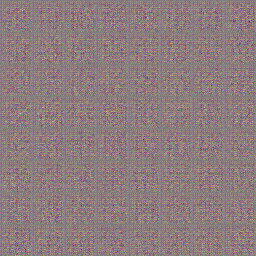}&
\includegraphics [width=0.17\linewidth]{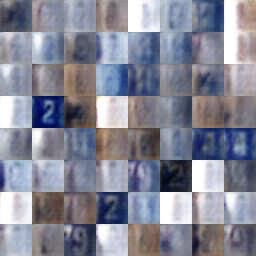}&
\includegraphics [width=0.17\linewidth]{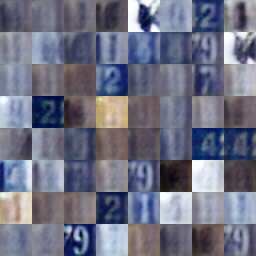}& 
\includegraphics [width=0.17\linewidth]{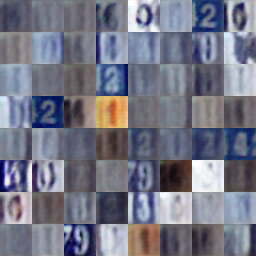}\\&
\includegraphics [width=0.17\linewidth]{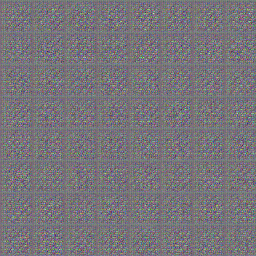}&
\includegraphics [width=0.17\linewidth]{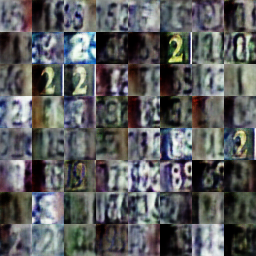}&
\includegraphics [width=0.17\linewidth]{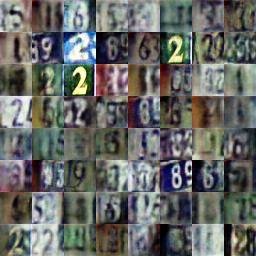} &
\includegraphics [width=0.17\linewidth]{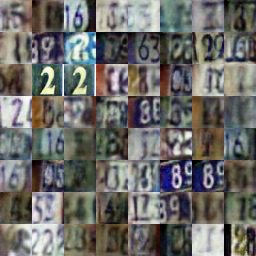} \\
\tiny{Mixed images}&\tiny{$1^{st}$ epoch} & \tiny{$6^{th}$ epoch} & \tiny{$17^{th}$ epoch} & \tiny{$39^{th}$ epoch}
\end{tabular}
\caption{\small{Evolution of output samples by two generators for fixed $z_1$ and $z_2$. The mixed images comprise two arbitrary digits drawn from $10$ digits in SVHN training dataset. The top generator outputs the samples, representing the lightning condition of the digits, while the bottom one generates samples from the distribution of all digits.}}
\label{SVHN_DemixingTraining_all}
\end{figure}

Now we present some experimental results with colorful images. Specifically, we use the demixing-GAN with SVHN dataset~\cite{netzer2011reading}. The Street View House Numbers (SVHN) training dataset is a collection of almost 70000 images, containing images of digits from 1 to 10. SVHN dataset is significantly more challenging to learn its distribution as it is noisy, including images of various resolution and distracting digits~\cite{chen2016infogan}. In our experiment, we use the character level representation of SVHN which are pre-processed $32\times32$ colorful images. Figure~\ref{SVHN_DemixingTraining_all} illustrates a similar experiment which we mentioned before. The mixed images also comprise the superposition of two randomly chosen digits from the SVHN training dataset. In addition, the dimension of the hidden space (i.e., $z$-space) for selecting random vectors $z_1$ and $z_2$ is set to 100. As expected, it takes more time compared to the other datasets for demixing-GAN to explore the distribution of the digits in SVHN. The authors of~\cite{chen2016infogan} pointed out that InfoGAN can capture two components in SVHN: Lighting and Context, where the context represents the central digit in an image. In demixing-GAN, we can see that one generator tries to learn the samples of SVHN dataset, the context part, while the other one mostly captures the lightning of the digits.

\subsection{Failure of the demixing-GAN}
\label{FailOfDemixing}
In this section, we empirically explore our observation about the failure of the demixing-GAN. As we discussed briefly in section~\ref{TestMnist}, we focus on two spaces, hidden space ($z$-space) and signal or generator space (the output of generator) in discovering the failure of demixing-GAN. 

\begin{figure}
\centering
\begin{tabular}{ccccc}
\multirow{1}{*}[2em]{\includegraphics [width=0.15\linewidth]{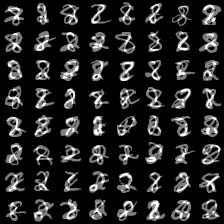}}&
\includegraphics [width=0.17\linewidth]{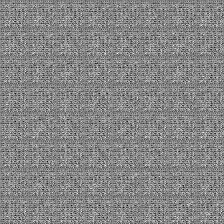}&
\includegraphics [width=0.17\linewidth]{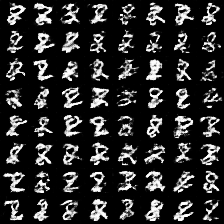}&
\includegraphics [width=0.17\linewidth]{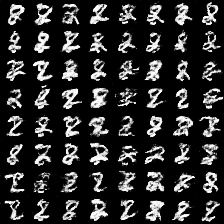}& 
\includegraphics [width=0.17\linewidth]{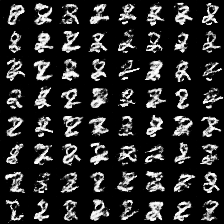}\\&
\includegraphics [width=0.17\linewidth]{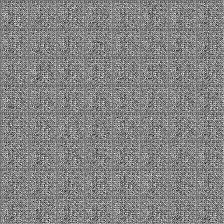}&
\includegraphics [width=0.17\linewidth]{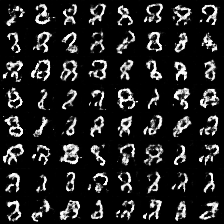}&
\includegraphics [width=0.17\linewidth]{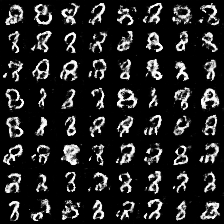} &
\includegraphics [width=0.17\linewidth]{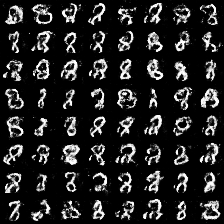} \\
\tiny{Mixed images}&\tiny{$1^{st}$ epoch} & \tiny{$6^{th}$ epoch} & \tiny{$15^{th}$ epoch} & \tiny{$21^{th}$ epoch}
\end{tabular}
\caption{\small{Failure of the demixing. Evolution of output samples by two generators for $z_1 = z_2$. The mixed images are the superposition of digits $2$ and $8$.}}
\label{fail_two_same_z}
\end{figure}

\begin{figure}
\centering
\begin{tabular}{ccccc}
\multirow{1}{*}[2em]{\includegraphics [width=0.15\linewidth]{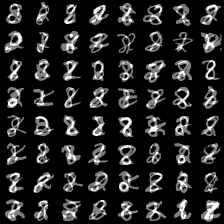}}&
\includegraphics [width=0.17\linewidth]{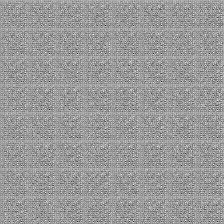}&
\includegraphics [width=0.17\linewidth]{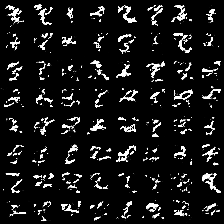}&
\includegraphics [width=0.17\linewidth]{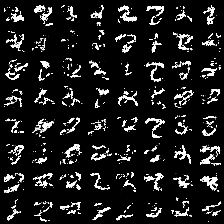}& 
\includegraphics [width=0.17\linewidth]{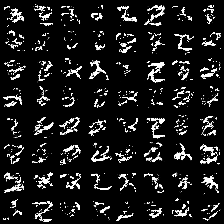}\\&
\includegraphics [width=0.17\linewidth]{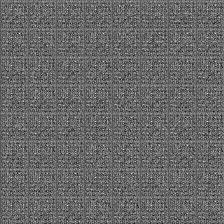}&
\includegraphics [width=0.17\linewidth]{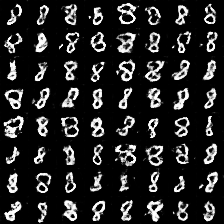}&
\includegraphics [width=0.17\linewidth]{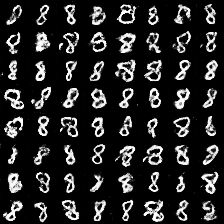} &
\includegraphics [width=0.17\linewidth]{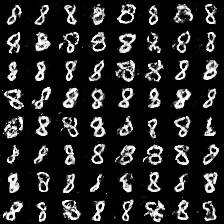} \\
\tiny{Mixed images}&\tiny{$1^{st}$ epoch} & \tiny{$6^{th}$ epoch} & \tiny{$15^{th}$ epoch} & \tiny{$21^{th}$ epoch}
\end{tabular}
\caption{\small{Failure of the demixing. Evolution of output samples by two generators for $z_1 = 0.1z_2$. The mixed images are the superposition of digits $2$ and $8$.}}
\label{fail_two_aligned_z}
\end{figure}

Our first observation concerns the $z$-space. We observe that if the hidden vectors form $z$-space of two generators are aligned to each other, then the two generators cannot output the samples in the signal space, representing the distribution of the constituent components. To be more precise, in Figure~\ref{fail_two_same_z}, we consider separating digits $8$ and $2$ from their superpositions similar to the experiment in the bottom panel of Figure~\ref{digits_DemixingTraining}. However, here, we feed both generators with the same vector, i.e., $z_1 = z_2$ in each batch (this is considered as the extreme case where precisely the hidden variables equal to each other) and track the evolution of the output samples generated by both generators. As we can see even after $21$ epochs, the generated samples by both generators are an unclear combination of both digits $2$ and $8$, and they are not separated clearly as opposed to the case when we feed the generators with $i.i.d$ random vectors. We also repeat the same experiment with two aligned vectors $z_1$ and $z_2$, i.e., $z_2= 0.1z_1$, Figure~\ref{fail_two_aligned_z} shows the evolution of the output samples generated by both generators for this setup. As shown in this experiment, two generators cannot learn the distribution of digits $8$ and $2$. While we do not currently have a mathematical argument for this observation, we conjecture that the hidden space ($z$-space) is one of the essential pieces in the demixing performance of the proposed demixing-GAN. We think that having (random) independent or close orthogonal vector $z$'s for the input of each generator is a necessary condition for the success of learning of the constituent components distribution, and consequently demixing of them. Further investigation of this line of study is indeed an interesting research direction, and we defer it for future research.
 
\begin{figure}[htb]
\centering
\begin{tabular}{ccccc}
{\includegraphics [width=0.17\linewidth]{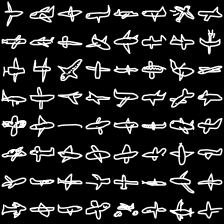}}&
\includegraphics [width=0.17\linewidth]{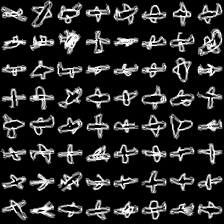}&
\includegraphics [width=0.17\linewidth]{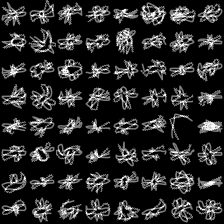}&
\includegraphics [width=0.17\linewidth]{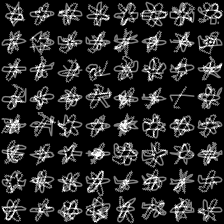}& 
\includegraphics [width=0.17\linewidth]{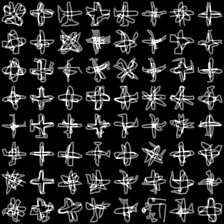}\\
{(a)}&{(b)} & {(c)} & {(d)} & {(e)}
\end{tabular}
\caption{\small{Mixed images of airplanes with different orientation, (a). Mixture of two $0^{\circ}$ rotated images (b). Mixture of $0^{\circ}$ and $10^{\circ}$ rotated images (c). Mixture of $0^{\circ}$ and $30^{\circ}$ rotated images rotated images (d). Mixture of $0^{\circ}$ and $60^{\circ}$ rotated images rotated images (e). Mixture of $0^{\circ}$ and $90^{\circ}$ rotated images rotated images.}}
\label{fail_two_airplane_rotations}
\end{figure}

In addition to the hidden space, here we design some experiments in the generator space that reveals the condition under which the demixing is failed. In particular, we consider the airplane images in Quick-Draw dataset. To construct the input mixed images, we consider randomly chosen images of the airplane from 16000 images as the first component. Then, the second component is constructed by rotating exactly the same one in the first components in a counterclockwise direction. We consider 5 different rotations, $0^{\circ}$, $10^{\circ}$, $30^{\circ}$, $60^{\circ}$, $90^{\circ}$. Five samples of such images are depicted in Figure~\ref{fail_two_airplane_rotations}. This experiment is sort of similar to the one in Figure~\ref{QD_DemixingTraining_twoair} in which we have seen that demixing-GAN can capture the internal structure in the airplane dataset by clustering it into two types.

\begin{figure}
\centering
\begin{tabular}{ccccc}
\multirow{1}{*}[2em]{\includegraphics [width=0.13\linewidth]{Two_qd_r0r0.png}}&
\includegraphics [width=0.11\linewidth]{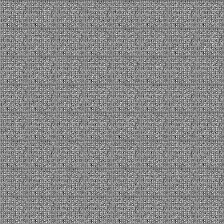}&
\includegraphics [width=0.11\linewidth]{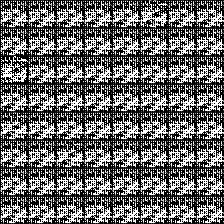}&
\includegraphics [width=0.11\linewidth]{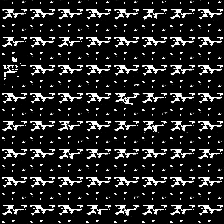}& 
\includegraphics [width=0.11\linewidth]{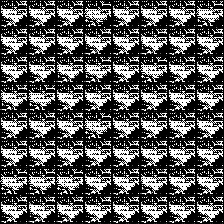}\\&
\includegraphics [width=0.11\linewidth]{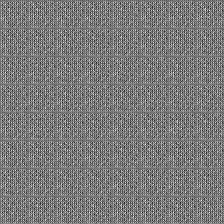}&
\includegraphics [width=0.11\linewidth]{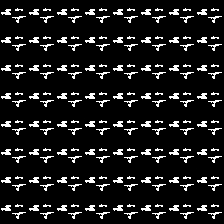}&
\includegraphics [width=0.11\linewidth]{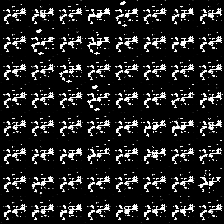} &
\includegraphics [width=0.11\linewidth]{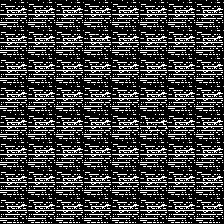} \\
{$0^{\circ}$-rotation}&\tiny{$1^{st}$ epoch} & \tiny{$5^{th}$ epoch} & \tiny{$13^{th}$ epoch} & \tiny{$31^{th}$ epoch}\\
\\
\multirow{1}{*}[2em]{\includegraphics [width=0.13\linewidth]{Two_qd_r0r10.png}}&
\includegraphics [width=0.11\linewidth]{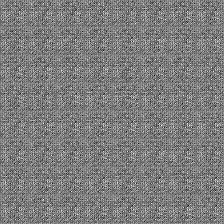}&
\includegraphics [width=0.11\linewidth]{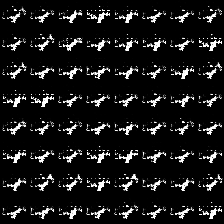}&
\includegraphics [width=0.11\linewidth]{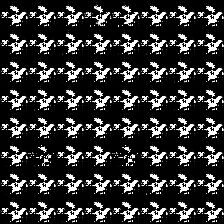}& 
\includegraphics [width=0.11\linewidth]{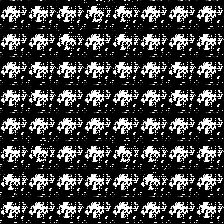}\\&
\includegraphics [width=0.11\linewidth]{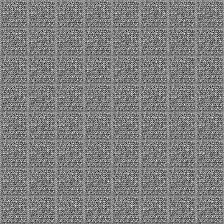}&
\includegraphics [width=0.11\linewidth]{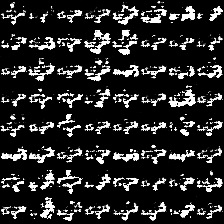}&
\includegraphics [width=0.11\linewidth]{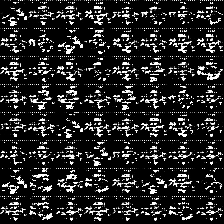} &
\includegraphics [width=0.11\linewidth]{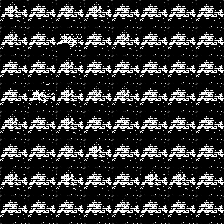} \\
{$10^{\circ}$-rotation}&\tiny{$1^{st}$ epoch} & \tiny{$5^{th}$ epoch} & \tiny{$13^{th}$ epoch} & \tiny{$31^{th}$ epoch}\\
\\
\multirow{1}{*}[2em]{\includegraphics [width=0.13\linewidth]{Two_qd_r0r30.png}}&
\includegraphics [width=0.11\linewidth]{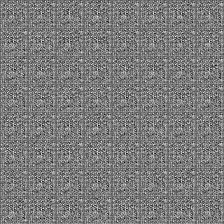}&
\includegraphics [width=0.11\linewidth]{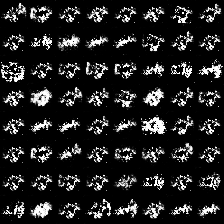}&
\includegraphics [width=0.11\linewidth]{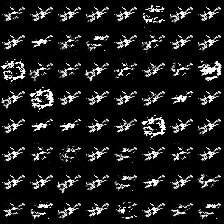}& 
\includegraphics [width=0.11\linewidth]{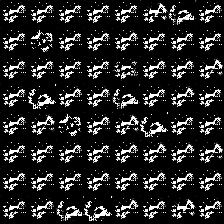}\\&
\includegraphics [width=0.11\linewidth]{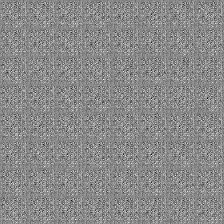}&
\includegraphics [width=0.11\linewidth]{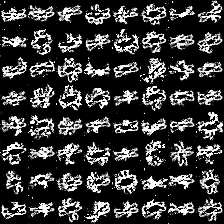}&
\includegraphics [width=0.11\linewidth]{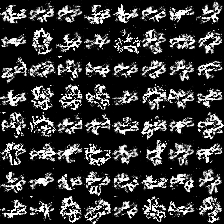} &
\includegraphics [width=0.11\linewidth]{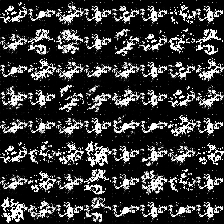} \\
{$30^{\circ}$-rotation}&\tiny{$1^{st}$ epoch} & \tiny{$5^{th}$ epoch} & \tiny{$13^{th}$ epoch} & \tiny{$31^{th}$ epoch}\\
\\
\multirow{1}{*}[2em]{\includegraphics [width=0.13\linewidth]{Two_qd_r0r60.png}}&
\includegraphics [width=0.11\linewidth]{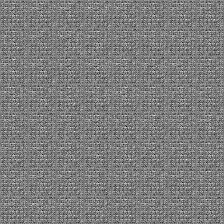}&
\includegraphics [width=0.11\linewidth]{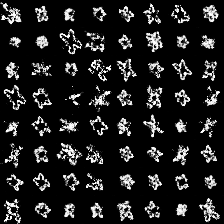}&
\includegraphics [width=0.11\linewidth]{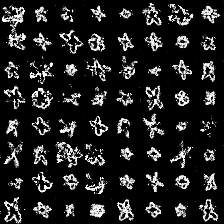}& 
\includegraphics [width=0.11\linewidth]{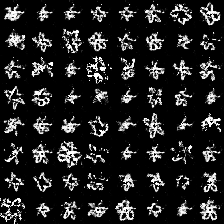}\\&
\includegraphics [width=0.11\linewidth]{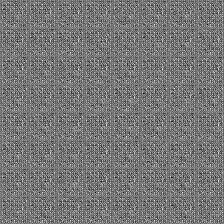}&
\includegraphics [width=0.11\linewidth]{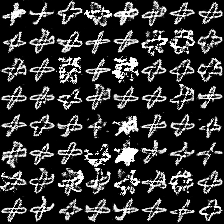}&
\includegraphics [width=0.11\linewidth]{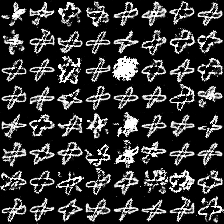} &
\includegraphics [width=0.11\linewidth]{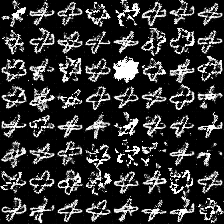} \\
{$60^{\circ}$-rotation}&\tiny{$1^{st}$ epoch} & \tiny{$5^{th}$ epoch} & \tiny{$13^{th}$ epoch} & \tiny{$31^{th}$ epoch}\\
\\
\multirow{1}{*}[2em]{\includegraphics [width=0.13\linewidth]{Two_qd_r0r90.png}}&
\includegraphics [width=0.11\linewidth]{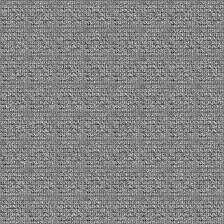}&
\includegraphics [width=0.11\linewidth]{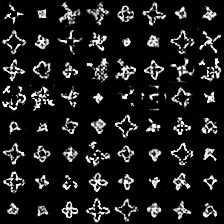}&
\includegraphics [width=0.11\linewidth]{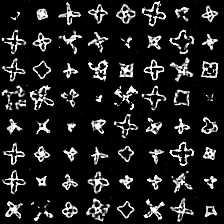}& 
\includegraphics [width=0.11\linewidth]{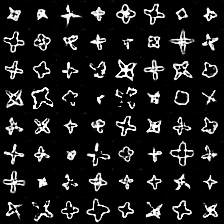}\\&
\includegraphics [width=0.11\linewidth]{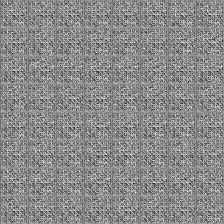}&
\includegraphics [width=0.11\linewidth]{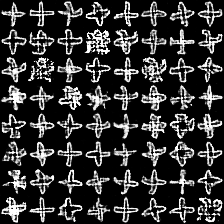}&
\includegraphics [width=0.11\linewidth]{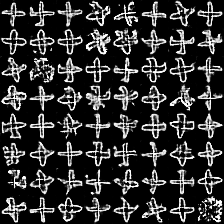} &
\includegraphics [width=0.11\linewidth]{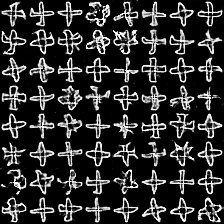} \\
{$90^{\circ}$-rotation}&\tiny{$1^{st}$ epoch} & \tiny{$5^{th}$ epoch} & \tiny{$13^{th}$ epoch} & \tiny{$31^{th}$ epoch}
\end{tabular}
\caption{\small{Failure of the demixing. Evolution of output samples by two generators. \textbf{Top:} Mixture of two $0^{\circ}$ rotated images. \textbf{Second Top:} Mixture of $0^{\circ}$ and $10^{\circ}$ rotated images. \textbf{Third Top:} Mixture of $0^{\circ}$ and $30^{\circ}$ rotated images rotated images. \textbf{Fourth Top:} Mixture of $0^{\circ}$ and $60^{\circ}$ rotated images rotated images. \textbf{Bottom:} Mixture of $0^{\circ}$ and $90^{\circ}$ rotated images rotated images.}}
\label{fail_two_diff_rotated_images}
\end{figure}

Now we perform the demixing-GAN on these datasets. Figure~\ref{fail_two_diff_rotated_images} illustrated the the evolution of the generators for various rotation degrees. The top panel shows the case exactly both components are the same shape. Obviously, the demixing, in this case, is impossible as there is no hope to distinguish the components from each other. Going down in the figure\ref{fail_two_airplane_rotations}, we have different rotation settings. As we can see, once we move forward to the $90^{\circ}$, both generators can capture the samples from the airplane distribution; however, as not clear as the case in which we had added the airplane shapes randomly for the input mixed images. We conjecture that changing the orientation of one component can make it incoherent to some extent from the other component, and consequently makes the demixing possible. In other words, we see again when two images show some distinguishable structures (in this case, the first one has $0$-oriented object and the other is the same one but rotated $90^{\circ}$), then the demixing-GAN can capture these structures.

